\documentclass[11pt]{article}

\usepackage[preprint]{acl}

\usepackage{times}
\usepackage{latexsym}

\usepackage[T1]{fontenc}

\usepackage[utf8]{inputenc}

\usepackage{microtype}

\usepackage{inconsolata}

\usepackage{graphicx}

\usepackage{multirow}
\usepackage{booktabs}
\usepackage{amsmath}
\usepackage{amsfonts}
\usepackage{subcaption}
\usepackage{threeparttable}
\usepackage{algorithm}
\usepackage{algorithmic}
\usepackage{makecell}

\usepackage{colortbl}
\usepackage{xcolor}
\newcommand{\red}[1]{\textcolor{red}{#1}}
\newcommand{\blue}[1]{\textcolor{blue}{#1}}

\usepackage[most]{tcolorbox}
\definecolor{colframecolor}{RGB}{55,98,175}   
\definecolor{colbackcolor}{RGB}{218,227,243}  

\usepackage{pifont}
\newcommand{\cmark}{\textcolor{green!70!black}{\ding{51}}} 
\newcommand{\xmark}{\textcolor{red}{\ding{55}}}            

\usepackage[misc]{ifsym}

\title{Think Outside the Policy: In-Context Steered Policy Optimization}

\author{
    Hsiu-Yuan Huang$^{1,2,3}$\thanks{~~Equal contribution. \fontsize{8.2pt}{8pt}\selectfont Work done during internship at Tencent.},
    Chenming Tang$^{1,2}$\footnotemark[1],\\
    \textbf{Weijie Liu}$^{3}$\thanks{~~Corresponding author.},
    \textbf{Clive Bai}$^{3}$,
    \textbf{Saiyong Yang$^{3}$,}
    \textbf{Yunfang Wu$^{1,2}$\footnotemark[2]}\\
    $^{1}$National Key Laboratory for Multimedia Information Processing, Peking University \\ 
    $^{2}$School of Computer Science, Peking University \quad
    $^{3}$LLM Department, Tencent \\
    \small{\textbf{Correspondence:}
    \href{mailto:hsiuyuanhuang.pku@gmail.com}{huang.hsiuyuan@stu.pku.edu.cn}\ \ \href{mailto:wuyf@pku.edu.cn}{wuyf@pku.edu.cn} }\\
}

\begin{document}
\maketitle

\begin{abstract}

Existing Reinforcement Learning from Verifiable Rewards (RLVR) methods, such as Group Relative Policy Optimization (GRPO), have achieved remarkable progress in improving the reasoning capabilities of Large Reasoning Models (LRMs). However, they exhibit limited exploration due to reliance on on-policy rollouts which are confined to the current policy's distribution, resulting in narrow trajectory diversity. Recent approaches attempt to expand policy coverage by incorporating trajectories generated from stronger expert models, yet this reliance increases computational cost and such advanced models are often inaccessible. To address these issues, we propose In-Context Steered Policy Optimization (ICPO), a unified framework that leverages the inherent in-context learning capability of LRMs to provide expert guidance using existing datasets. ICPO introduces mixed-policy GRPO with implicit expert forcing, which expands exploration beyond the current policy distribution without requiring advanced LRM trajectories. To further stabilize optimization, ICPO integrates expert region reject sampling to filter unreliable off-policy trajectories and annealed expert-bonus reward shaping to balance early expert guidance with later autonomous improvement.
Results demonstrate that ICPO consistently enhances RLVR performance and training stability on mathematical reasoning benchmarks, revealing a scalable and effective RLVR paradigm for LRMs. Our code is available at \url{https://github.com/Celine-hxy/ICPO}.

\end{abstract}
\section{Introduction}

\begin{figure}[t]
\centering
  \includegraphics[width=0.95\columnwidth]{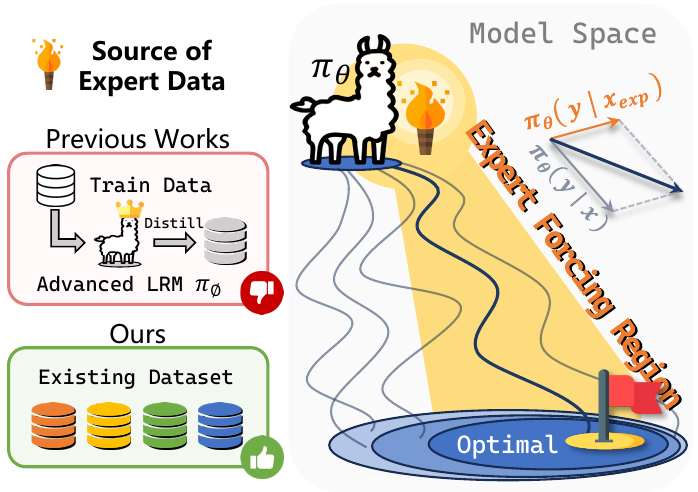}
  \caption{Illustration of optimization dynamics in parameter space. GRPO exploration is confined to the current policy’s distribution, limiting trajectory diversity and often leading to suboptimal convergence. While prior methods expand exploration by incorporating expert rollouts generated by stronger LRMs on the training data, ICPO leverages existing datasets as in-context guidance, eliminating reliance on advanced LRMs.}
  \label{fig:teaser}
\end{figure}

Large Reasoning Models (LRMs) excel at solving complex mathematical problems, and Reinforcement Learning from Verifiable Rewards (RLVR) provides a scalable way to refine their reasoning through verifiable rewards \cite{deepseekR12025}. Yet, limited exploration under standard Group Relative Policy Optimization (GRPO)~\cite{grpo2024deepseek}  often hinders robust reasoning improvement \cite{yue2025doesreinforcementlearningreally, zhang2025stephintmultilevelstepwisehints}.

Recent work has explored combining Supervised Fine-Tuning (SFT) with Reinforcement Learning (RL) to strengthen the exploration of LRMs.
One research direction interleaves SFT and RL updates \cite{ma2025ReLIFT}, enabling SFT to improve high-difficulty problem-solving while RL refines mid- and low-difficulty behaviors. However, repeated switching between paradigms introduces instability and inefficient convergence.
Another line of work seeks to unify SFT and RL within a single training process through different strategies: incorporating SFT data as off-policy rollouts during RL to expand the exploration space~\cite{luffy2025}; jointly optimizing SFT and RL objectives for tighter integration~\cite{fu2025SRFT}; and leveraging hints to bootstrap rollouts to improve performance on harder prompts~\cite{liu2025UFT,zhang2025BREAD,fu2025SRFT}.

Together, these approaches reflect a growing consensus that combining SFT and RL can effectively expand the exploration of policy search into more promising reasoning spaces. However, as shown in Figure \ref{fig:teaser}, three key challenges remain:
(1) Online exploration remains confined to the current policy distribution, as GRPO-based methods rely on on-policy sampling, resulting in limited trajectory diversity and might converge to local optima.
(2) Expanding the exploration space with trajectories from stronger LRMs incurs high computational cost and limited accessibility, since generating additional reasoning traces from advanced models is expensive and such models are not always available.
(3) External trajectories are often noisy and unstable for training, incorporating them indiscriminately as off-policy rollouts can mislead policy updates and harm convergence stability.

To address these limitations, we propose \textit{\textbf{I}n-\textbf{C}ontext Steered \textbf{P}olicy \textbf{O}ptimization} (ICPO), a novel RLVR framework that exploits the LRM’s inherent In-Context Learning (ICL) capability to provide expert guidance instead of relying on external advanced LRMs. Specifically, 
(1) ICPO introduces \textit{mixed-policy GRPO with Implicit Expert Forcing} (IEF), where expert-conditioned rollouts are generated through ICL guidance, enabling exploration beyond the current policy distribution and steering the model toward expert-aligned regions of the solution space.
(2) To ensure reliable guidance, ICPO employs \textit{Expert Region Reject Sampling} (ERRS), which filters out noisy or low-quality off-policy trajectories using verifiable reward signals, preventing misleading gradients from contaminating policy updates.
(3) ICPO further explores an annealed expert bonus into the \textit{Reward Shaping} (RS) design, enforcing strong expert-guided shaping in the early stage and progressively relaxing it to facilitate autonomous optimization as LRM capabilities grow.

Our experiments show that ICL-based rollouts exhibit superior quality and diversity, making them effective sources of expert guidance (\S~\ref{sec:preliminary}). Extensive evaluations across multiple mathematical reasoning benchmarks further demonstrate that ICPO achieves substantial performance gains over vanilla GRPO and mixed-policy GRPO~\cite{luffy2025}, with maximum average improvements of up to \textbf{+4.1} and \textbf{+4.0} points, respectively (\S~\ref{sec:result}). In addition, we provide in-depth instance- and token-level analyses that offer strong evidence of distributional shifts induced by ICPO (\S~\ref{sec:distribution_analysis}).

Our contributions are three-fold:
\begin{itemize}
    \item We empirically show that ICL rollouts provide diverse and high-quality expert signals for the mathematical reasoning task.
    \item We introduce ICPO, which includes \textit{mixed-policy GRPO with Implicit Expert Forcing}, leveraging the model’s inherent ICL ability without relying on stronger LRMs, together with \textit{Expert Region Reject Sampling} and \textit{Reward Shaping} for stable and efficient learning.
    \item We empirically validate that ICPO delivers consistent improvements on mathematical reasoning benchmarks across model scales and effectively encourages exploration, demonstrating strong potential for LRM post-training.
\end{itemize}

\section{Related Work}

\paragraph{Exploration from Within the Policy.}
Recent methods enhance exploration by increasing diversity within the policy itself, such as generating additional rollouts~\cite{wang2025eframe, hu2025brorlscalingRL, tang2025steprivertwicelearning}, leveraging replay buffers to revisit informative prompts~\cite{dou2025replay}, and sampling under higher temperatures to boost stochasticity~\cite{chen2025flaminghot, zhang2024edtimprovinglargelanguage}.
These approaches broaden trajectory coverage under the model’s own distribution. However, their exploration remains bounded by the intrinsic capacity of the current policy.

\paragraph{Exploration from Outside the Policy.}
Other studies leverage external supervision to guide exploration beyond the model’s default policy distribution.
One line of work explores hybrid SFT–RL strategies to expand the reasoning space of LRMs. ReLIFT \cite{ma2025ReLIFT} alternates between RL and SFT by updating on failed rollouts. LUFFY \cite{luffy2025} incorporates SFT trajectories as off-policy samples using importance sampling. SRFT \cite{fu2025SRFT} jointly optimizes SFT and RL objectives with an entropy-based weight on the SFT loss.
Another line of work guides rollouts by concatenating partial SFT solutions as hints \cite{liu2025UFT,zhang2025BREAD,huang2025RLPrefixSampling}.
However, these methods rely on advanced LRMs to supply SFT traces, which may incur additional computation overhead.

In contrast, we exploit LRMs' inherent ICL ability to steer diverse rollouts with existing datasets as demonstrations, requiring neither external LRMs’ trajectories nor explicitly engineered hints.

\section{Preliminary}
\label{sec:preliminary}
\paragraph{Explicit Expert Forcing.}
In traditional RL and RLHF, \textit{expert forcing} explicitly constrains the policy to align with an expert policy $\pi_\phi$, typically through imitation or KL-based regularization \cite{2018DeepQFromDemons, haldar23Imitation, zhang2023policy, hu2023imitation}.
This explicit constraint stabilizes optimization and reduces reward variance, but it requires gradient-based imitation and access to an auxiliary expert model (e.g., a larger LRM), which can limit exploration in later stages.

\begin{figure}[t]
\centering
  \includegraphics[width=\columnwidth]{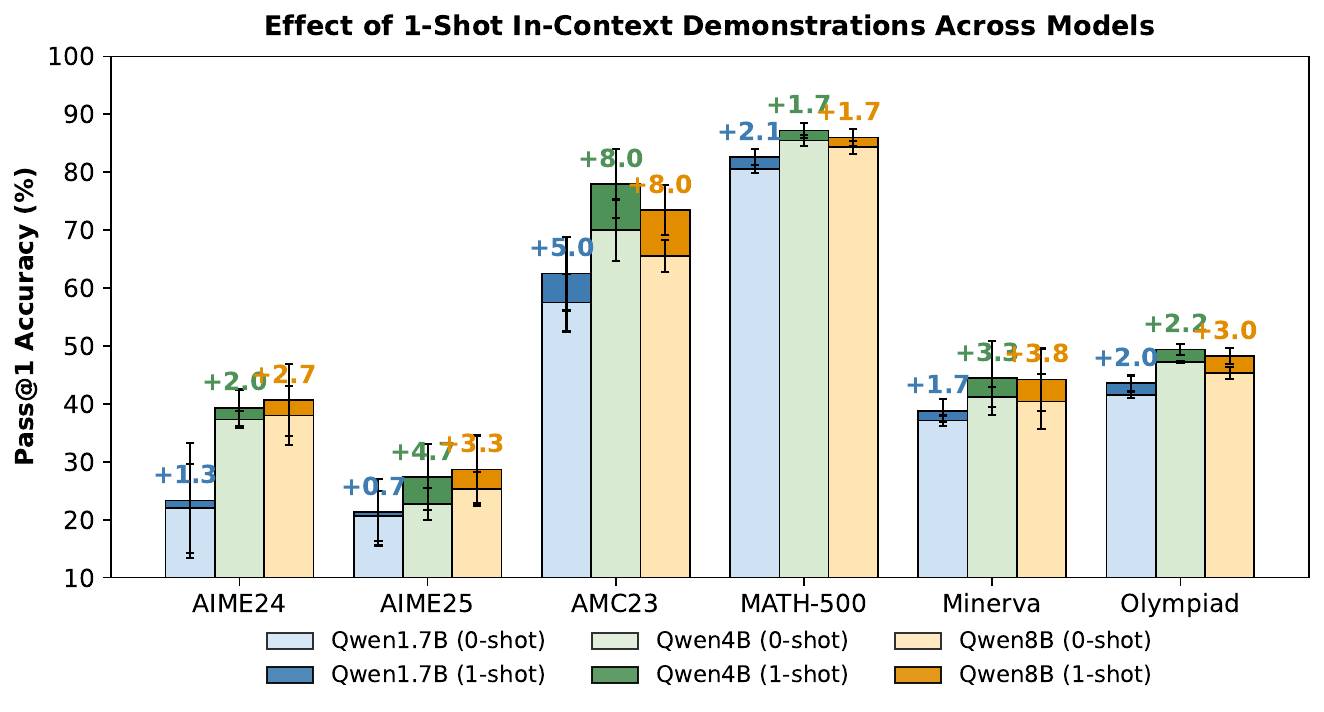}
  \caption{Comparison between 0-shot and 1-shot ICL on reasoning accuracy across benchmark datasets.}
  \label{fig:shot0vs1}
\end{figure}

\begin{figure}[t]
  \centering
  \includegraphics[width=\columnwidth]{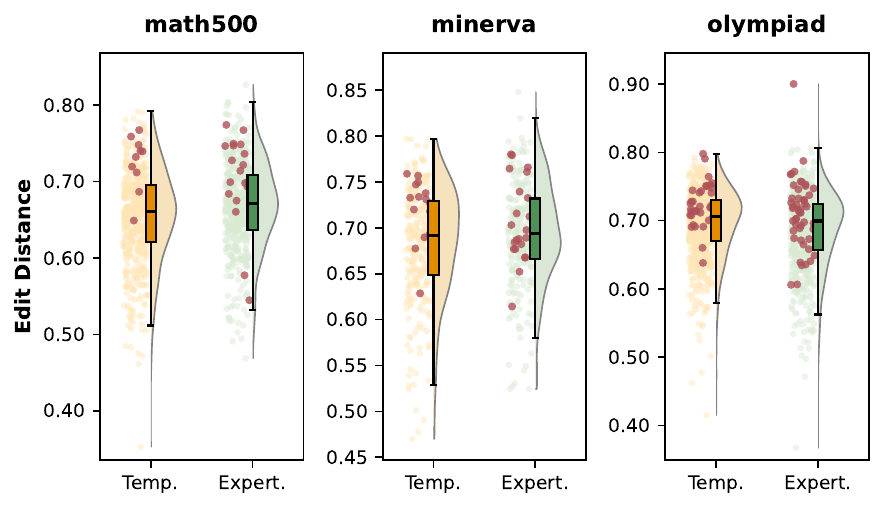}
  \caption{Effect of in-context steering on exploration and diversity. \textit{Temp.} increases the default decoding temperature (0.6→1.2), and \textit{Expert.} introduces in-context guidance (0-shot→1-shot). Red dots indicate instances flipped from incorrect to correct.}
  \label{fig:steering_effect}
\end{figure}

\paragraph{ICL as Expert-Conditioned Inference.}
ICL provides a gradient-free way to inject expert priors through the input context, which we evaluate from three complementary perspectives.  
(1)~\textit{Accuracy}: the 1-shot ICL setting consistently outperforms the 0-shot baseline (Figure~\ref{fig:shot0vs1}), indicating that conditioning on demonstrations improves reasoning correctness.  
(2)~\textit{Diversity}: compared with temperature-based sampling, introducing 1-shot demonstrations expands the sampling space (Figure~\ref{fig:steering_effect}), yielding larger inter-trajectory edit distances and enhanced exploratory diversity. 
(3)~\textit{Distribution quality}: under ICL-conditioned rollouts, the output distribution becomes more favorable—a higher proportion of previously incorrect generations are ``flipped'' to correct solutions compared with temperature perturbations (Figure~\ref{fig:steering_effect}), indicating that in-context steering provides a stronger and more targeted exploration signal.
Taken together, these results support our view that ICL constitutes an effective expert-conditioned inference process.

\paragraph{From ICL to Implicit Expert Forcing.}
Given expert demonstrations $\mathcal{D}$ and a query $q$, the model generates trajectories conditioned on:
\begin{equation}
x_{\mathrm{exp}} = [\mathcal{D}; q], \quad
\tau_{\mathrm{exp}} \sim \pi_\theta(\tau \mid x_{\mathrm{exp}}).
\end{equation}
Following the \textit{hypothesis-class} view of ICL~\cite{hendel2023taskvector}, the forward process of a Transformer $T$ can be decomposed into two functions:
\begin{equation}
T([\mathcal{D}, q]) = \mathcal{F}(q; A(\mathcal{D})),
\end{equation}
where $A(\cdot)$ maps demonstrations $\mathcal{D}$ to a task vector $\vartheta = A(\mathcal{D})$ that encodes the expert behavior specific to that task~\cite{hendel2023taskvector, NEURIPS2023_81b83900, todd2024functionvectorslargelanguage, huang2024multimodal, liu2024incontextvectorsmakingcontext, hojel2024findingvisualtaskvectors,
cai-2025-dyvec}, 
and $\mathcal{F}(\cdot\,; \vartheta)$ represents the task-specific reasoning function that applies the $\vartheta$ to generate the prediction for query $q$.
This leads to the parametric representation:
\begin{equation}
\pi_\theta^{\text{IEF}}(\tau \mid q)
:= \pi_\theta(\tau \mid [\mathcal{D}; q])
= \pi_\mathcal{F}(\tau \mid q; \vartheta),
\end{equation}
indicating that ICL implicitly introduces an \textit{expert-induced prior} $\vartheta$ that steers the rollout distribution toward expert-like regions—without any explicit optimization on $\pi_\theta$.
While ICL itself is an inference-time mechanism that does not update model parameters, we incorporate its induced trajectories into GRPO training to form IEF.

\paragraph{Group Relative Policy Optimization (GRPO).}
GRPO is an efficient \textit{On-Policy} optimization algorithm tailored for RL in LLMs, where the advantage for each token is computed in a group-relative manner without requiring an additional critic model to estimate token values. Given a set of rollouts $\{\tau_i\}_{i=1}^N$ sampled from the old policy $\pi_{\theta_{\text{old}}}$, the normalized advantage is computed by:
\begin{equation}\small
A_i = \frac{R(\tau_i) - \mathrm{mean}(G)}{\mathrm{std}(G)}, \quad 
G = \{R(\tau_i)\}_{i=1}^N.
\label{eq:grpo_adv}
\end{equation}
Analogous to PPO~\cite{schulman2017ppo}, the GRPO objective is formulated as:
\begin{equation}\small
\mathcal{J}_{\scriptscriptstyle \text{GRPO}}(\theta)
 = \frac{1}{\sum_{i=1}^N |\tau_i|} \sum_{i=1}^N \sum_{t=1}^{|\tau_i|} \text{CLIP}(r_{i,t}(\theta), A_i, \epsilon),
\end{equation}
where
$r_{i,t}(\theta) =
\frac{\pi_\theta(\tau_{i,t} | \tau_{i,<t})}
{\pi_{\theta_{\text{old}}}(\tau_{i,t} | \tau_{i,<t})}$
is the importance ratio, and
$\text{CLIP}(r, A, \epsilon) =
\min(r \cdot A, \text{clip}(r; 1-\epsilon, 1+\epsilon) \cdot A)$
is the clipping function for variance reduction.
To prevent the learned policy from drifting too far from the reference model, we retain the KL regularization term $\beta \cdot D_{\text{KL}}[\pi_\theta || \pi_{\text{ref}}]$ in GRPO, which is jointly optimized to ensure training stability and maintain controllable policy updates.

By leveraging ICL-conditioned rollouts within a mixed-policy GRPO framework, our approach enables expert-guided exploration to directly participate in policy optimization, effectively realizing an \textit{In-Context Steered Policy Optimization} process.

\section{Method}

\begin{figure*}[t]
\centering
  \includegraphics[width=0.95\linewidth]{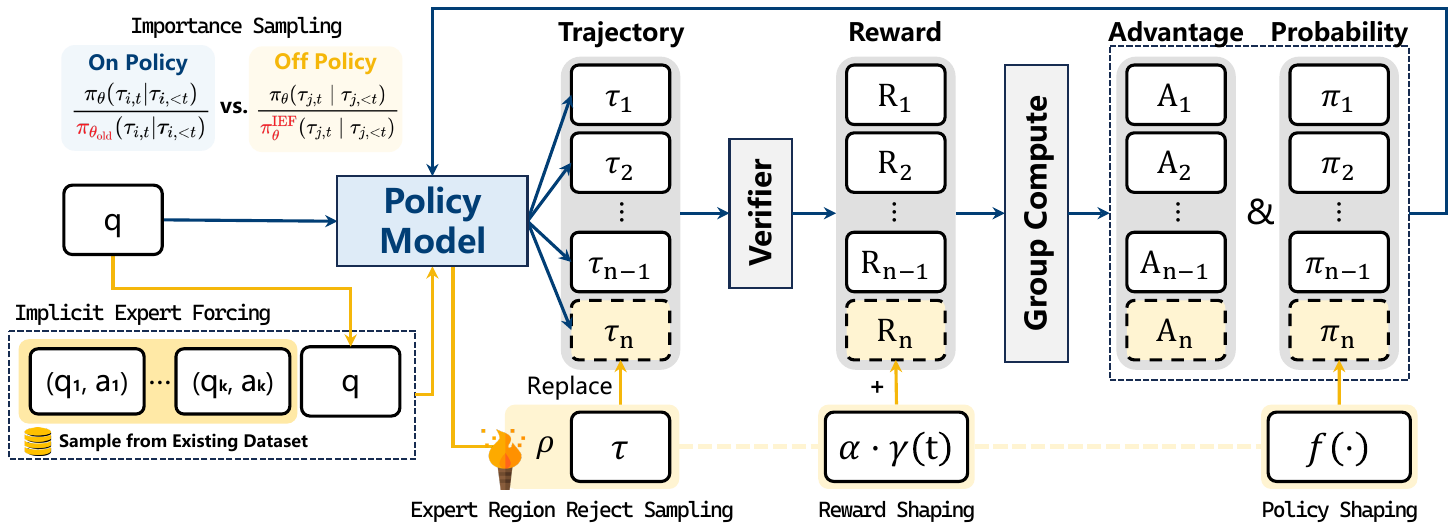}
  \caption{Overall Framework of ICPO. ICPO performs mixed-policy GRPO using off-policy trajectories generated by the policy model itself via implicit expert forcing, with reject sampling and reward shaping to stabilize training.}
  \label{fig:main}
\end{figure*}

Figure~\ref{fig:main} illustrates the overall ICPO training framework and the ICPO training process is detailed in Algorithm~\ref{alg:icpo}.

\subsection{Mixed-Policy GRPO with Implicit Expert Forcing}
To incorporate expert-conditioned exploration into group rollouts, we follow \citet{luffy2025} and extend GRPO into a \textit{mixed-policy} setting, where each group consists of $N_{\text{on}}$ on-policy trajectories $\tau_i \sim \pi_{\theta_{\text{old}}}$ and $N_{\text{off}}$ trajectories generate under IEF $\tau_j \sim \pi_{\theta_{\text{old}}}^{\text{IEF}}$, such that $N_{\text{on}}+N_{\text{off}}=N$.
We can recompute the group-normalized advantage (as in Eq.~\ref{eq:grpo_adv}) over the mixed rollout set as:
\begin{equation} \small
\hat{A}_i = \frac{R(\tau_i) - \text{mean}(G_{\text{on}} \cup G_{\text{off}})}{\text{std}(G_{\text{on}} \cup G_{\text{off}})},
\label{eq:mixAdv}
\end{equation}
where $G_{\text{on}} = \{R(\tau_i)\}_{i=1}^{N_{\text{on}}}$ and $G_{\text{off}} = \{R(\tau_j)\}_{j=1}^{N_{\text{off}}}$.

The objective of mixed-policy GRPO with IEF balances exploitation of the current policy with exploration toward expert-aligned regions, and can be written as follows:
\begin{equation}\small
\begin{aligned}
\mathcal{J}_{\scriptscriptstyle \text{Mixed}}(\theta)
&=
\underbrace{\mathbb{E}_{\tau \sim \pi_{\theta_{\text{old}}}}}_{\text{on-policy}}
\Bigg[
\frac{1}{|\tau|}\sum_{t=1a}^{|\tau|}
\operatorname{CLIP}\big(r_{t}(\theta),\, \hat{A}(\tau),\, \epsilon\big)
\Bigg]
\\
&\quad+
\underbrace{\mathbb{E}_{\tau \sim \pi_{\theta_{\text{old}}}^{\text{IEF}}}}_{\text{off-policy}}
\Bigg[
\frac{1}{|\tau|}\sum_{t=1}^{|\tau|}
\operatorname{CLIP}\big(\hat r_{t}(\theta),\, \hat{A}(\tau),\, \epsilon\big)
\Bigg],
\end{aligned}
\label{eq:mix_grpo_objective}
\end{equation}
where $\hat{r}_{j,t}(\theta)=\frac{\pi_\theta(\tau_{j,t}\mid\tau_{j,<t})}{\pi^{\text{IEF}}_\theta(\tau_{j,t}\mid\tau_{j,<t})}$ is the expert-conditioned importance weight.

Unlike prior work~\cite{luffy2025}, which adopts a \textit{model-based off-policy} scheme by relying on an additional advanced LRM $\pi_\phi$ to provide expert trajectories for the same training prompts, our mixed-policy GRPO with IEF operates as an \textit{input-conditioned off-policy} method.
Specifically, while all rollouts are sampled from the same policy $\pi_\theta$, in-context demonstrations alter the input conditioning and steer the policy away from its default output distribution. 
This conditioning discrepancy induces a behavior mismatch, under which expert-conditioned rollouts $\tau_j \sim \pi_\theta(x_{\text{exp}})$ are regarded as off-policy relative to standard on-policy samples $\tau_i \sim \pi_\theta(x)$.

\subsection{Expert Region Reject Sampling}
Building upon the expert-conditioned off-policy branch above, we further restrict updates to those trajectories that demonstrably improve model performance. 
We define an \textit{expert region} as the subset of states where expert conditioning yields superior guidance, steering the policy beyond its native distribution. A rollout $\tau_j$ generated under expert conditioning is accepted into this region if its reward exceeds a predefined threshold $\delta$:
\begin{equation}
\mathcal{E}_{\text{exp}} = 
\big\{ (x_{\text{exp}}, \tau_j) 
\ \big| \ R(\tau_j) \ge \delta \big\},
\end{equation}
where $\delta$ is set to $1.0$ by default.

To prevent low-quality expert-conditioned traces from biasing training, we define a reject sampling operator $\rho$ that selectively retains trajectories within the expert region. 
Formally, $\rho$ performs reject sampling by restricting the expectation to trajectories that fall within the expert region:
\begin{equation}
\rho(\cdot) =
\mathbb{E}_{\tau \sim \pi_{\theta}(\tau \mid \tau \in \mathcal{E}_{\text{exp}})}
[g(\tau)],
\end{equation}
where $g(\tau)$ denotes the per-trajectory contribution to the objective.
This filtering ensures that only high-reward expert-conditioned rollouts contribute to policy updates. 
The final objective of ICPO then becomes:
\begin{equation}\small
\begin{aligned}
\mathcal{J}_{\mathrm{ICPO}}(\theta)=\frac{1}{Z}\Big[
\underbrace{\sum_{i=1}^{N_{\mathrm{on}}}\sum_{t=1}^{|\tau_{i}|}
\mathrm{CLIP}(r_{i,t}(\theta),A_{i},\epsilon)}_{\text{on-policy objective}} \\
+ \underbrace{\rho\left(\sum_{j=1}^{N_{\mathrm{off}}}\sum_{t=1}^{|\tau_{j}|}
\mathrm{CLIP}(f(\hat{r}_{j,t}(\theta)),\hat{A}_{j},\epsilon)\right)}_{\text{off-policy objective}}
\Big],
\end{aligned}
\label{eq:icpo_objective}
\end{equation}
where $Z$ normalizes over all valid tokens.
The shaping function $f(\cdot)$ follows prior work and is defined as $f(x)=\frac{x}{x+\lambda}$, where $\lambda=0.01$ by default~\cite{luffy2025}. 
This shaping biases learning toward expert-induced improvements while encouraging exploration.

\subsection{Reward Shaping with Annealed Expert Bonus}
The verifiable reward function evaluates the model output by extracting the final answer and comparing it against the predefined ground-truth answer. 
It assigns a binary score based on whether the extracted answer matches the correct solution under a task-specific verifier. 
Formally,
\begin{equation}
R(\tau)={\left\{\begin{array}{l l}{1}&{{\text{if}}\ \tau\ {\text{is correct}}}\\ {0}&{{\text{otherwise.}}}\end{array}\right.}
\end{equation}
This verifiable reward has been shown to reliably lead to successful scaling of RL training.

To encourage early imitation of expert-conditioned behavior while avoiding long-term over-reliance, we design a variant of ICPO (namely ICPO$\dagger$) and add a step-annealed bonus only to trajectories that have the correct answer and within the expert region $\mathcal{E}_{\text{exp}}$:
\begin{equation}
R_{\text{shaped}}(\tau) 
= R(\tau) + \alpha \cdot \gamma(t),
\end{equation}
where $\gamma(t) = 1 - \frac{t}{T}$ denotes a linear decay scheduler over the training step $t$, and $\alpha$ denotes the bonus weight (set to $1.0$ in our experiments).

\begin{algorithm}[t]
\small
\caption{ICPO Training Procedure}
\label{alg:icpo}
\begin{algorithmic}[1]
\REQUIRE Policy $\pi_{\theta}$, old policy $\pi_{\theta_{\text{old}}}$, 
expert data $\mathcal{D}$, batch size $B$, rollout size $N$,
few-shot count $k$, RS threshold $\delta$, step $t$, annealed bonus $\alpha \cdot \gamma(t)$
\FOR{each step}
    \STATE Sample prompts $\{x_i\}_{i=1}^B$
    \FOR{$i = 1$ to $B$}
        \FOR{$j = 1$ to $N$}
            \STATE $\tau_i^j \sim \pi_{\theta_{\text{old}}}(\cdot|x_i)$
            \STATE Compute $R(\tau_i^j)$
        \ENDFOR
        \STATE Sample $k$ expert (q, a) pairs from $\mathcal{D}$ and form $x_i^{\text{exp}}$
        \STATE Generate $\tau_i^{\text{IEF}} \sim \pi_{\theta_{\text{old}}}^{\text{IEF}}(\cdot|x_i^{\text{exp}})$
        \IF{$R(\tau_i^{\text{IEF}}) \ge \delta$ \AND $\mathrm{correct}(\tau_i^{\text{IEF}})$}
            \STATE Pick random $j$
            \STATE Replace $\tau_i^j \leftarrow \tau_i^{\text{IEF}}$
            \IF{Enable reward shaping}
                \STATE $R(\tau_i^j) \leftarrow R(\tau_i^{\text{IEF}}) + \alpha \cdot \gamma(t)$
            \ENDIF
        \ENDIF
        \STATE Compute $\hat{A}_i$ using Eq.~\ref{eq:mixAdv}
    \ENDFOR
    \STATE Compute mixed rollout loss $\mathcal{L}$ according to $\mathcal{J}_{\mathrm{ICPO}}(\theta)$\\
    \STATE $\theta \leftarrow \theta - \eta \nabla_\theta \mathcal{L}$
    \STATE $\pi_{\theta_{\text{old}}} \leftarrow \pi_\theta$
\ENDFOR
\end{algorithmic}
\end{algorithm}

\section{Experimental Setup}

\paragraph{Dataset.} We follow \citet{luffy2025} and adopt the \textit{OpenR1-Math-220k} dataset as our main training corpus. 
Specifically, we use the filtered subset\footnote{\scriptsize \href{https://huggingface.co/datasets/Elliott/Openr1-Math-46k-8192}{\texttt{https://huggingface.co/datasets/Elliott/Openr1-Math-46k}}}, which excludes generations exceeding 8192 tokens as well as those identified as incorrect by \textit{Math-Verify}\footnote{\scriptsize \url{https://github.com/huggingface/Math-Verify}}.
The resulting dataset contains approximately 45k verified reasoning prompts.
For IEF, instead of using trajectories from advanced LRMs, we randomly sample demonstrations for each prompt from the \textit{MATH}~\cite{math2021} training set, which contains 7.5k mathematical problems paired with high-quality solutions.

\paragraph{Implementation Details.}
The complete set of hyperparameters used in our training is listed in Appendix \ref{sec:train_setting}. 
For the main experiments, we use \textit{Qwen3-1.7B} and \textit{Qwen3-8B} \cite{yang2025qwen3technicalreport} as the base models and employ GRPO~\cite{grpo2024deepseek} as our RL algorithm. 
We also include \textit{Qwen2.5-Math-7B} to facilitate comparison with other related methods. 
We additionally evaluate ICPO on \textit{Qwen3-8B-Base} and \textit{LLaMA-3.1-8B} to examine its cross-model generalization.
We generate a total of 8 rollout trajectories per prompt. For the on-policy baseline, we use 8 on-policy rollouts. For our mixed-policy GRPO, we follow previous work~\cite{luffy2025} and use 1 off-policy rollout and 7 on-policy rollouts to ensure comparability.

\paragraph{Evaluation Settings.}
\label{sec:eval_setting}
We evaluate on six widely used mathematical reasoning benchmarks: \textbf{AIME24}, \textbf{AIME25}, \textbf{AMC23}~\cite{li2024numinamath}, \textbf{Minerva}~\cite{Minerva2022}, \textbf{Olympiad}~\cite{he2024olympiadbench}, and \textbf{MATH-500} \cite{math2021}. 
To assess generalization beyond in-domain reasoning, we further test on three out-of-distribution (OOD) benchmarks: \textbf{ARC-C}~\cite{clark2018arc_c}, \textbf{GPQA-Diamond}~\cite{rein2023gpqa}, and \textbf{MMLU-Pro}~\cite{mmlu2024}, with multiple-choice options shuffled to prevent contamination. For main experiments, the AIME24, AIME25, and AMC23 benchmark, which have relatively small test sets, we report \textbf{\textit{Avg@32}} (for \textit{\textbf{Pass@1}}), while for the other benchmarks we report \textbf{\textit{Pass@1}}.
More evaluation details are provided in Appendix \ref{sec:eval_setting}.

\paragraph{Baseline Methods.}
For direct comparison, we evaluate our ICPO against the vanilla \textbf{GRPO} baseline trained on the same subset of \textit{OpenR1-Math-220k}.
We further include \textbf{$\text{GRPO}_{\text{ExtraRollouts}}$}, which extends GRPO by using 16 on-policy rollouts to enhance exploration.
To examine the effect of training data source, we also train a GRPO variant on expert-domain data, identical to our IEF demonstrations (\textit{MATH}), denoted as \textbf{$\text{GRPO}_{\text{ExpertDomain}}$}.
In addition, we compare with \textbf{LUFFY}~\cite{luffy2025}, which leverages trajectories generated by advanced LRMs as off-policy rollouts, along with other baselines detailed in Appendix~\ref{sec:baselines}.

\section{Experimental Results}
\label{sec:result}
\subsection{Main Results}

\begin{table*}[htbp]
\centering
\small
\setlength{\tabcolsep}{2pt}
\begin{tabular}{lcccccc@{\hskip 0pt}c|cccc@{\hskip 0pt}c}
\toprule
& \multicolumn{7}{c}{\textbf{In-Distribution Benchmarks}} 
& \multicolumn{5}{c}{\textbf{Out-of-Distribution Benchmarks}} \\
\cmidrule(lr){2-8}
\cmidrule(lr){9-13}
\textbf{Method}
& \textbf{AIME 24/25} 
& \textbf{AMC23} 
& \textbf{MATH} 
& \textbf{Minerva} & \textbf{Olympiad}
& \textbf{Avg.} & \textbf{(Impr.)}
& \textbf{ARC} & \textbf{GPQA} & \textbf{MMLU}
& \textbf{Avg.} & \textbf{(Impr.)}\\
\midrule

\textbf{\textit{Qwen3-1.7B}}
& 21.7 / 20.6 & 56.8 & 79.4 & 37.1 & 40.6 
& 42.7 & --
& 88.3 & 22.2 & 52.3 & 54.3 & --\\

\midrule
\; GRPO
& 28.4 / 22.5 & 66.7 & 83.6 & 40.8 & 48.2 
& 48.4 & --
& 88.3 & 34.3 & 54.4 & 59.0 & --\\

\; $\text{GRPO}_{\text{ExtraRollouts}}$
& 28.3 / 24.7 & 69.8 & 84.4 & 43.0 & 53.8 
& 50.7 & \red{(+2.3)}
& \textbf{88.9} & 30.8 & 55.0 & 58.2 & \blue{(-0.8)}\\

\; $\text{GRPO}_{\text{ExpertDomain}}$
& 26.8 / 24.8 & 66.3 & 83.8 & \textbf{45.6} & 50.5 
& 49.6 & \red{(+1.2)}
& 88.8 & 32.8 & \textbf{55.5} & 59.0 & \blue{(0.0)}\\

\midrule
\; ICPO (Ours)
& \textbf{31.3} / 26.3 & \textbf{70.4} & 86.8 & 44.1 & \textbf{56.4}
& \textbf{52.5} & \textbf{\red{(+4.1)}}
& 88.1 & 27.8 & \textbf{55.5} & 57.1 & \blue{(-1.9)}\\

\; ICPO$\dagger$ (Ours)
& 29.0 / \textbf{26.6} & 70.0 & \textbf{87.2} & 42.7 & 52.7
& 51.4 & \red{(+3.0)}
& 87.7 & \textbf{36.4} & 55.0 & \textbf{59.7} & \red{(+0.7)}\\

\midrule\midrule

\textbf{\textit{Qwen3-8B}}
& 33.7 / 21.4 & 65.4 & 86.2 & 40.4 & 43.3 
& 48.4 & --
& 96.2 & 37.9 & 68.0 & 67.4 & --\\

\midrule
\; GRPO
& 54.8 / 38.5 & 83.8 & 91.0 & 50.7 & 62.4
& 63.5 & --
& 95.8 & 51.0 & 72.0 & 72.9 & --\\

\; $\text{GRPO}_{\text{ExtraRollouts}}$
& 53.0 / 40.0 & 85.4 & \textbf{93.0} & \textbf{52.9} & 61.6
& 64.3 & \red{(+0.8)}
& 95.7 & 53.0 & 72.9 & 73.9 & \red{(+1.0)}\\

\; $\text{GRPO}_{\text{ExpertDomain}}$
& \textbf{58.3} / 40.4 & 86.2 & 91.8 & 50.7 & 60.0
& 64.6 & \red{(+1.1)}
& 92.2 & 51.0 & 70.2 & 71.1 & \blue{(-1.8)}\\

\midrule
\; ICPO (Ours)
& 55.2 / \textbf{43.7} & \textbf{87.0} & 92.0 & 51.1 & \textbf{65.2}
& \textbf{65.7} & \textbf{\red{(+2.2)}}
& 95.5 & \textbf{55.1} & 72.3 & 74.3 & \red{(+1.4)}\\

\; ICPO$\dagger$ (Ours)
& 56.2 / 40.9 & 85.4 & 92.0 & 51.5 & 64.3
& 65.0 & \red{(+1.5)}
& \textbf{95.6} & \textbf{55.1} & \textbf{75.3} & \textbf{75.3} & \textbf{\red{(+2.4)}}\\

\bottomrule
\end{tabular}
\caption{Evaluation results for Qwen3. ICPO shows clear and consistent improvements on in-distribution data, while ICPO$\dagger$ provides a more balanced trade-off between in-distribution and OOD performance.}
\label{tab:main_results}
\end{table*}

\begin{table*}[t]
\centering
\small
\setlength{\tabcolsep}{3pt}
\begin{tabular}{lcccccccc}
\toprule
\textbf{Method} 
& \textbf{AIME24} 
& \textbf{AIME25} 
& \textbf{AMC23} 
& \textbf{MATH} 
& \textbf{Minerva} 
& \textbf{Olympiad} 
& \textbf{Average} \\
\midrule
\textbf{\textit{Qwen2.5-Math-7B}}~\cite{yang2025qwen3technicalreport}
& 11.5 & 4.9  & 31.3 & 43.6 & 7.4 & 15.6 & 19.0 \\
\midrule
\multicolumn{8}{l}{\textbf{Previous RLVR Methods}}\\[3pt]
\; SimpleRL-Zero*~\cite{zeng2025simplerl}           
& 27.0 & 6.8  & 54.9 & 76.0 & 25.0 & 34.7 & 37.4 \\
\; PRIME-Zero*~\cite{cui2025PRIMEzero}              
& 17.0 & 12.8 & 54.0 & 81.4 & 39.0 & 40.3 & 40.8 \\
\; OpenReasoner-Zero*~\cite{hu2025openreasonerzero} 
& 16.5 & 15.0 & 52.1 & 82.4 & 33.1 & 47.1 & 41.0 \\
\; Oat-Zero*~\cite{liu2025oatzero}                  
& 33.4 & 11.9 & 61.2 & 78.0 & 34.6 & 43.4 & 43.8 \\

\midrule
\multicolumn{8}{l}{\textbf{SFT and RL}}\\[3pt]
\; SFT~\cite{luffy2025}
& 23.8 & 22.9 & 61.8 & 82.2 & 37.9 & 42.1 & 45.1 \\
\; RL$_{\text{GRPO}}$~\cite{grpo2024deepseek}
& 22.9 & 13.3 & 63.0 & 81.2 & 37.1 & 43.1 & 43.4 \\
\; SFT+RL$_{\text{GRPO}}$~\cite{grpo2024deepseek}
& 29.3 & 22.3 & 67.1 & 85.8 & 44.1 & 50.8 & 49.9 \\

\midrule
\multicolumn{8}{l}{\textbf{Comparable Baseline Methods}}\\[3pt]
\; ReLIFT*~\cite{ma2025ReLIFT}   
& 28.2 & 22.9 & 64.8 & 85.0 & 37.1 & 54.9 & 48.8 \\
\; LUFFY*~\cite{luffy2025}
& 29.4 & 23.1 & 65.6 & 87.6 & 37.5 & \textbf{57.2} & 50.1 \\
\; Prefix-RFT*~\cite{huang2025RLPrefixSampling}
& 31.8 & 26.4 & 68.2 & \textbf{88.4} & 40.3 & 55.7 & 51.8 \\
[1pt]
\rowcolor{black!5}
\; ICPO (Ours)
& 29.3 & \textbf{28.9} & 74.9 & \textbf{88.4} & \textbf{45.6} & 52.4 & 53.2 \\
\rowcolor{black!5}
\; ICPO$\dagger$ (Ours)
& \textbf{32.8} & 26.7 & \textbf{75.9} & 86.6 & 44.9 & 53.6 & \textbf{53.4} \\
\bottomrule
\end{tabular}
\caption{Evaluation results for Qwen2.5-Math-7B. “*” indicate results reported from their original papers.}
\label{tab:qwen25_in_dis_result}
\end{table*}

\begin{figure*}[t]
  \centering
  \begin{subfigure}[t]{0.32\linewidth}
    \centering
    \includegraphics[width=\linewidth]{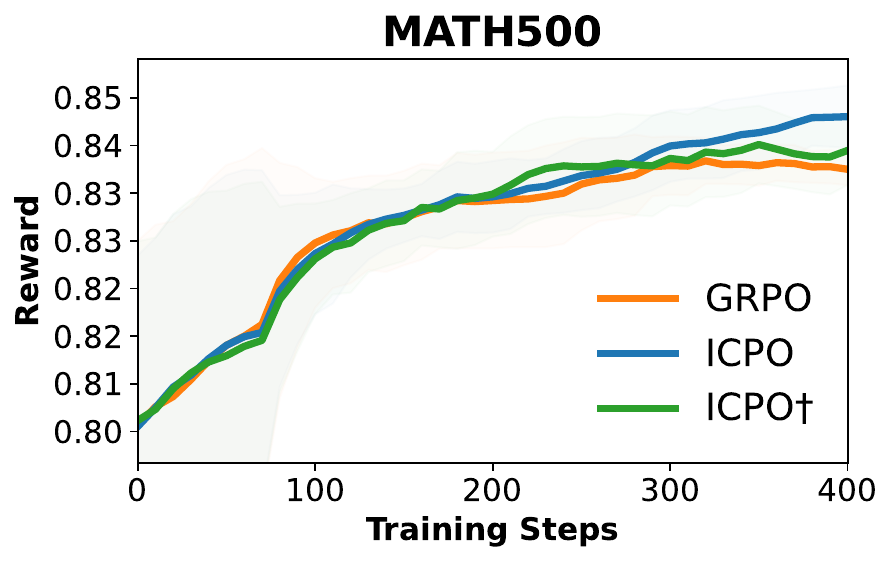}
    \caption{Expert Domain Reward}
  \end{subfigure}
  \begin{subfigure}[t]{0.32\linewidth}
    \centering
    \includegraphics[width=\linewidth]{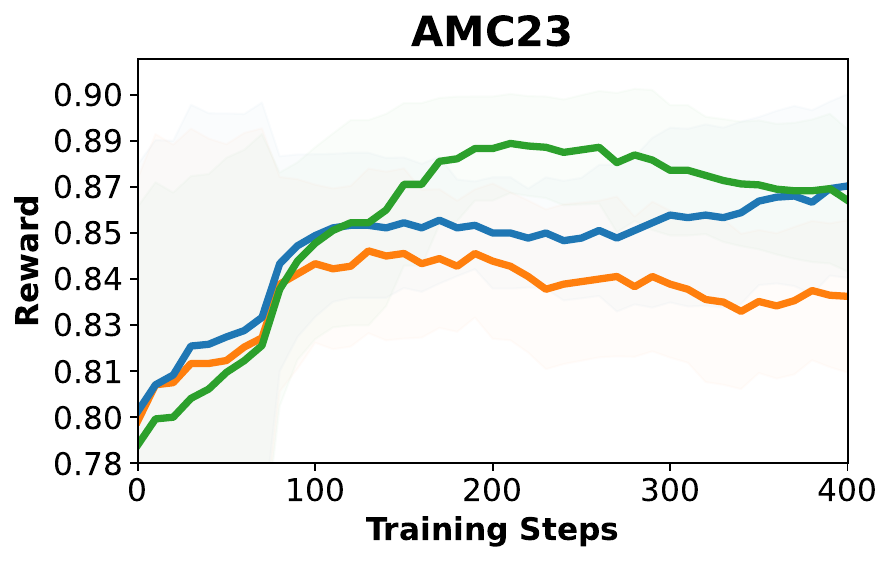}
    \caption{In-Distribution Reward}
  \end{subfigure}
  \begin{subfigure}[t]{0.32\linewidth}
    \centering
    \includegraphics[width=\linewidth]{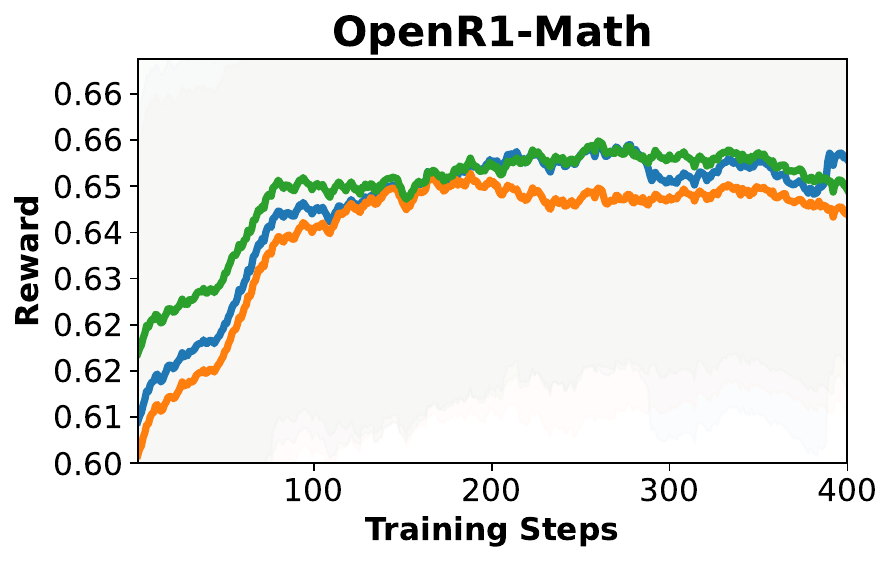}
    \caption{Training Reward}
  \end{subfigure}
  \caption{Reward curves over training steps across test (\textit{Mean@2}) and train sets.}
  \label{fig:reward}
\end{figure*}

\begin{figure*}[t]
\centering
\begin{subfigure}[t]{0.31\linewidth}
  \centering
  \includegraphics[width=\linewidth]{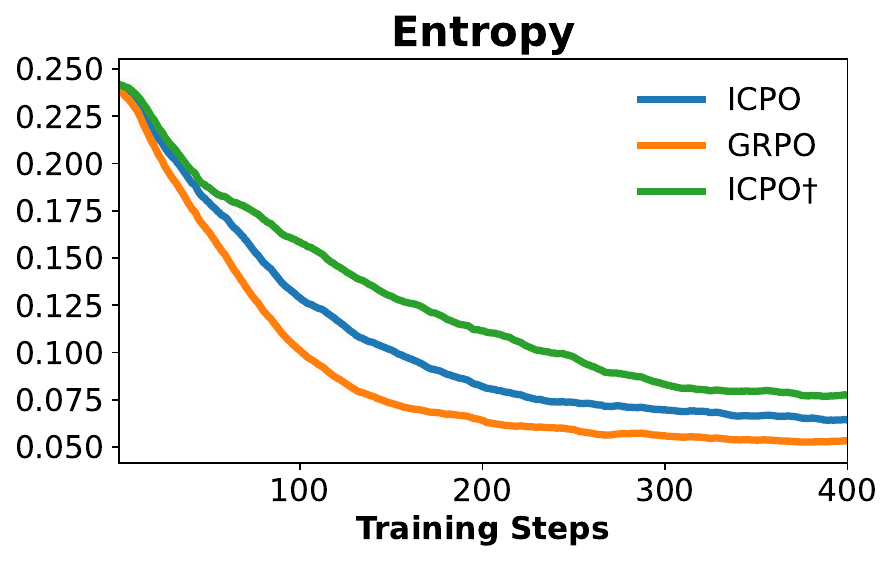}
\end{subfigure}
\begin{subfigure}[t]{0.31\linewidth}
  \centering
  \includegraphics[width=\linewidth]{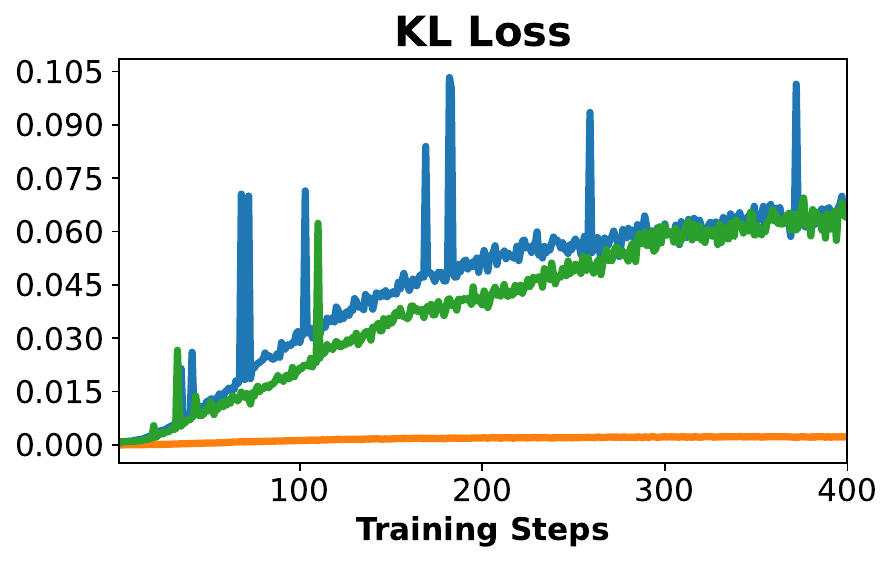}
\end{subfigure}
\begin{subfigure}[t]{0.31\linewidth}
  \centering
  \includegraphics[width=\linewidth]{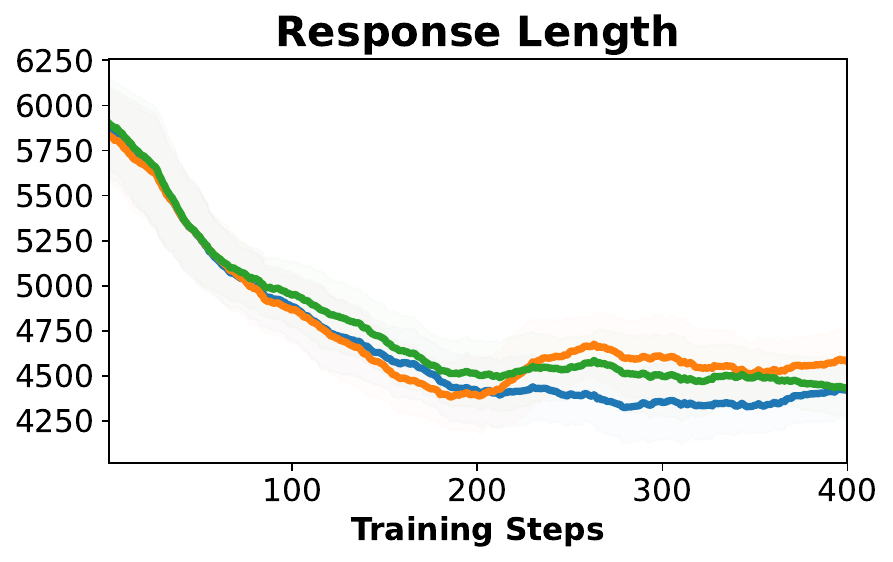}
\end{subfigure}

\caption{Training dynamics of Qwen3-1.7B.}
\label{fig:train_dynamic_1_7B}
\end{figure*}

The main experiments include two variants of our proposed ICPO framework: \textbf{ICPO}, which operates without reward shaping (RS), and \textbf{ICPO$\dagger$}, which incorporates RS to further enhance expert-domain alignment. 
To better understand their optimization behavior against GRPO, we visualize the reward dynamics over training steps across different datasets, as shown in Figure~\ref{fig:reward} and \ref{fig:reward_curves_1_7B}, where both ICPO variants consistently achieve higher rewards throughout training. We also visualize the training dynamics in Figure~\ref{fig:train_dynamic_1_7B}, where ICPO maintains a higher policy entropy than GRPO, reflecting a broader policy support and increased exploration during training.

\paragraph{In-Distribution Evaluation.} Table \ref{tab:main_results} shows both the in- and out-of-distribution performance, where \textit{MATH-500} serves as the expert domain.
Across both model scales, ICPO consistently outperforms the vanilla GRPO baseline, especially on in-distribution benchmarks.
For the smaller Qwen3-1.7B model, ICPO$\dagger$ achieves an average improvement of \textbf{+3.0} points over GRPO, while ICPO further stabilizes optimization with a \textbf{+4.1} point overall gain. A similar trend is observed for the larger Qwen3-8B model, where ICPO and ICPO$\dagger$ yield \textbf{+2.2} and \textbf{+1.5} average improvements, respectively.

These consistent gains across scales indicate that ICPO effectively steers the policy toward a more optimal output distribution.
By leveraging both the challenging training set and high-quality expert demonstrations through IEF, ICPO forms a more informative training signal than either source alone—a synergistic effect that yields superior performance over both $\text{GRPO}_{\text{ExpertDomain}}$ and vanilla GRPO.
Moreover, unlike ICPO, simply increasing the number of on-policy rollouts in $\text{GRPO}_{\text{ExtraRollouts}}$ offers only marginal improvement, revealing the inherent exploration limitations of vanilla GRPO and underscoring the importance of effective steering.

\paragraph{Out-of-Distribution Evaluation.}
ICPO$\dagger$ introduces RS to explicitly amplify the advantage of trajectories falling within the expert domain $\mathcal{E}_{\text{exp}}$.
Our hypothesis is that steering optimization toward these expert-aligned trajectories induces more coherent reasoning structures and stabilizes policy updates, which in turn enhance generalization both in- and out-of-distribution. Empirically, ICPO$\dagger$ achieves average gains of \textbf{+0.7} and \textbf{+2.4} over GRPO on Qwen3-1.7B and 8B, respectively, providing evidence supporting this hypothesis.

\paragraph{Comparison with Other Baselines.}
Table~\ref{tab:qwen25_in_dis_result} presents in-distribution results compared to existing baselines. ICPO$\dagger$ achieves the strongest performance among all comparable methods, surpassing ReLIFT~\cite{ma2025ReLIFT}, LUFFY~\cite{luffy2025}, and Prefix-RFT~\cite{huang2025RLPrefixSampling} by \textbf{+4.6}, \textbf{+3.3}, and \textbf{+1.6} points, respectively.
For brevity, the results and analysis on OOD benchmarks are provided in Appendix~\ref{sec:qwen25ood}.

\subsection{Ablation Study}

\paragraph{Effect of Each Component.}

\begin{table}[t]
\centering
\small
\begin{threeparttable}
\setlength{\tabcolsep}{2pt}
\begin{tabular}{lccccc}
\toprule
\textbf{Variant} & \textbf{MATH} & \textbf{AIME24/25} & \textbf{AMC} & \textbf{Mnrv.} & \textbf{Avg.} \\
\midrule
\multicolumn{6}{l}{\textbf{\textit{Qwen3-1.7B}}} \\ [3pt]
ICPO
& \textbf{86.8} & 31.3 / \textbf{26.3} & \textbf{70.4} & \textbf{44.1} & \textbf{51.8} \\
- ERRS      
& 85.6 & \textbf{32.2} / 25.9 & 66.8 & 42.3 & 50.6 \\
- IEF (GRPO) 
& 83.6 & 28.4 / 22.5 & 66.7 & 40.8 & 48.4 \\
\midrule
\multicolumn{6}{l}{\textbf{\textit{Qwen3-8B}}} \\ [3pt]
ICPO
& \textbf{92.0} & \textbf{55.2} / \textbf{43.6} & \textbf{87.0} & 51.1 & \textbf{65.8} \\
- ERRS      
& 89.6 & \textbf{55.2} / 41.7 & 85.2 & \textbf{53.3} & 65.0 \\
- IEF (GRPO) 
& 91.0 & 54.8 / 38.5 & 83.8 & 50.7 & 63.8 \\

\bottomrule
\end{tabular}
\vspace{0.25em}
{\scriptsize\raggedright
RS = \textit{Reward Shaping}, 
ERRS = \textit{Expert Region Reject Sampling}, IEF = \textit{Implicit Expert Forcing}.\par}
\caption{Ablation analysis by progressively removing components from ICPO.}
\label{tab:ablation_component}
\end{threeparttable}
\end{table}


Table~\ref{tab:ablation_component} summarizes ablations by progressively removing components of ICPO.
On Qwen3-8B, all components contribute positively: \textit{IEF} yields the largest gain (\textbf{+1.2}) by injecting expert-conditioned guidance and enhancing exploration, while \textit{ERRS} improves accuracy by filtering invalid expert-region trajectories (\textbf{+0.8}).
Across both model sizes, removing any component consistently harms performance, demonstrating that all the modules are complementary and jointly essential for the effectiveness of ICPO.

\paragraph{Selection of Expert Data.}
\begin{table}[t]
\centering
\small
\setlength{\tabcolsep}{3.5pt}
\begin{tabular}{lcccccc}
\toprule
\textbf{Method} 
& \textbf{MATH} 
& \textbf{AIME24/25} 
& \textbf{AMC} 
& \textbf{Mnrv.} 
& \textbf{Avg.} \\
\midrule
\textbf{\textit{Qwen3-1.7B}}\\[3pt]
GRPO
& 83.6 & 28.4 / 22.5 & 66.7 & 40.8 & 48.4 \\ [1pt]

ICPO (PoT)
& \textbf{87.0} & \textbf{31.5} / \textbf{27.0} & 69.6 & 42.3 & 51.5 \\
ICPO (CoT)
& 86.8 & 31.3 / 26.3 & \textbf{70.4} & \textbf{44.1} & \textbf{51.8} \\

\midrule
\textbf{\textit{Qwen3-8B}}\\[3pt]
GRPO
& 91.0 & 54.8 / 38.5 & 83.8 & 50.7 & 63.8 \\ [1pt]

ICPO (PoT)
& 90.8 & \textbf{56.7} / 40.7 & 83.8 & \textbf{53.7} & 65.1 \\
ICPO (CoT)
& \textbf{92.0} & 55.2 / \textbf{43.6} & \textbf{87.0} & 51.1 & \textbf{65.8} \\

\bottomrule
\end{tabular}
\caption{Ablation on expert data selection, where CoT refers to mathematical data (\textit{MATH}) and PoT refers to code-oriented OOD reasoning data.
}
\label{tab:ablation_expert_data}
\end{table}
To demonstrate the robustness of ICPO to the choice of expert data, we replace the in-domain expert data with cross-domain Program-of-Thought (PoT) data \cite{yue2023mammothPoT} and conduct ICPO training under this setting. Details of data processing are provided in Appendix~\ref{sec:ablation_expert_guidance}. As shown in Table~\ref{tab:ablation_expert_data}, ICPO consistently outperforms GRPO when using either CoT or PoT data as demonstrations for implicit expert guidance. These results suggest that ICPO is robust to the choice of expert data and highlight its potential to generalize beyond mathematical reasoning to cross-domain settings.
More importantly, by changing the expert data, ICPO functions as a \textbf{\textit{plug-and-play}} framework that flexibly reshapes the target policy distribution, enabling controllable steering of model behavior.

\paragraph{Sources of Expert Guidance.}

\begin{table}[t]
\centering
\small
\begin{threeparttable}
\setlength{\tabcolsep}{1.2pt}
\begin{tabular}{lc@{\hskip 0pt}c@{\hskip 0pt}cccccc}
\toprule
\textbf{Variant} & \textbf{A.} & \textbf{E.} & \textbf{R.} & \textbf{MATH} & \textbf{AIME24/25} & \textbf{AMC} & \textbf{Mnrv.} & \textbf{Avg.} \\
\midrule
\multicolumn{9}{l}{\textbf{\textit{Qwen3-1.7B}}} \\ [3pt]
LUFFY & \cmark & \xmark & \xmark & 83.6 & 26.8 / 23.5 & 63.3 & 41.9 & 47.8 \\ 
ICPO$^*$ & \xmark & \cmark & \xmark 
& 85.6 & \textbf{32.2} / 25.9 & 66.8 & 42.3 & 50.6 \\
ICPO & \xmark & \cmark & \cmark
& \textbf{86.8} & 31.3 / \textbf{26.3} & \textbf{70.4} & \textbf{44.1} & \textbf{51.8} \\

\midrule
\multicolumn{9}{l}{\textbf{\textit{Qwen3-8B}}} \\ [3pt]
LUFFY & \cmark & \xmark & \xmark & 91.0 & 53.1 / 37.0 & 85.3 & 52.2 & 63.7 \\ 
ICPO$^*$ & \xmark & \cmark & \xmark & 89.6 & \textbf{55.2} / 41.7 & 85.2 & \textbf{53.3} & 65.0 \\
ICPO & \xmark & \cmark & \cmark & \textbf{92.0} & \textbf{55.2 }/ \textbf{43.6} & \textbf{87.0} & 51.1 & \textbf{65.8} \\
\bottomrule
\end{tabular}
\vspace{0.25em}
{\scriptsize\raggedright
A. = \textit{Advanced LRM Trajectory}, E. = \textit{Existing Dataset}, 
R. = \textit{Expert Region Reject Sampling (ERRS)}. ICPO$^*$ = \textit{ICPO w/o ERRS}.\par}
\caption{
Comparison across expert guidance sources. 
}
\label{tab:icpo_luffy}
\end{threeparttable}
\end{table}

We compare ICPO with LUFFY~\cite{luffy2025}, which incorporates trajectories generated by advanced LRMs into off-policy GRPO, as shown in Table~\ref{tab:icpo_luffy}.
ICPO$^*$, which removes ERRS and is thus directly comparable to LUFFY in its source of expert guidance, already surpasses LUFFY.
This demonstrates that IEF can steer the model toward a better policy distribution by leveraging existing datasets as contextual guidance, eliminating the need for costly external LRM computation. Please refer to Appendix \ref{sec:ablation_appendix} for more ablation studies.

\paragraph{Role of Base Model ICL Capability.}
\begin{table}[htbp]
\centering
\small
\setlength{\tabcolsep}{2pt}
\begin{tabular}{lcccccc}
\toprule
\textbf{Method} 
& \textbf{MATH} 
& \textbf{AIME24/25} 
& \textbf{AMC} 
& \textbf{Mnrv.} 
& \textbf{Avg.} \\

\midrule

\textbf{\textit{Pass@1}} \\[3pt]
GRPO
& 84.6 & 22.5 / 21.4 & \textbf{71.8} & 43.8 & 48.8 \\

$\text{GRPO}_{\text{ExtraRollouts}}$
& 85.8 & \textbf{26.9} / 20.2 & 67.4 & 46.7 & 49.4 \\

$\text{GRPO}_{\text{ExpertDomain}}$
& 80.0 & 22.9 / 12.8 & 60.9 & \textbf{47.8} & 44.9 \\ [1pt]

\rowcolor{black!5}
ICPO (Ours)
& \textbf{88.6} & 26.8 / \textbf{21.6} & 69.2 & 46.3 & \textbf{50.5} \\


\midrule

\textbf{\textit{Pass@32}} \\[3pt]
GRPO
& \textbf{96.2} & 53.3 / 43.3 & 90.0 & 58.8 & 68.3 \\

$\text{GRPO}_{\text{ExtraRollouts}}$
& 95.4 & 50.0 / 46.7 & 92.5 & \textbf{64.3} & 69.8 \\

$\text{GRPO}_{\text{ExpertDomain}}$
& 92.0 & 46.7 / 26.7 & 90.0 & 61.0 & 63.3 \\ [1pt]

\rowcolor{black!5}
ICPO (Ours)
& 95.8 & \textbf{60.0} / \textbf{60.0} & \textbf{95.0} & 63.6 & \textbf{74.9} \\


\bottomrule
\end{tabular}
\caption{Pass@1 and Pass@32 results for \textit{Qwen3-8B-Base}, a model with no ICL capability.}
\label{tab:qwen3_8b_base_icl}
\end{table}
To examine whether ICPO relies on the assumption that the base model possesses basic ICL capability, we conduct additional experiments on \textit{Qwen3-8B-Base}, a model with no ICL capability. The results are summarized in Table~\ref{tab:qwen3_8b_base_icl}.
The observed \textit{Pass@1} improvement demonstrates that ICPO remains effective even when the base model lacks ICL ability. Meanwhile, the substantial \textit{Pass@32} improvement indicates that ICPO expands the exploration space rather than merely sharpening existing modes, encouraging the model to “think outside the policy.” These results show that ICPO is not predicated on the ICL capabilities of the base model Instead, ICL acts as a scaling amplifier: stronger ICL ability leads to larger improvements, yet weakly aligned models can also benefit from implicit expert forcing during RL optimization.

\section{Analysis of Distribution Shift}
\label{sec:distribution_analysis}
To examine how ICPO alters the model’s predictive distribution, we conduct an in-depth analysis on the expert-domain dataset \textit{MATH-500} from both the instance- and token-level perspectives.

\begin{figure}[t]
  \centering
  \begin{subfigure}[t]{0.49\linewidth}
    \centering
    \includegraphics[width=\linewidth,height=\linewidth]{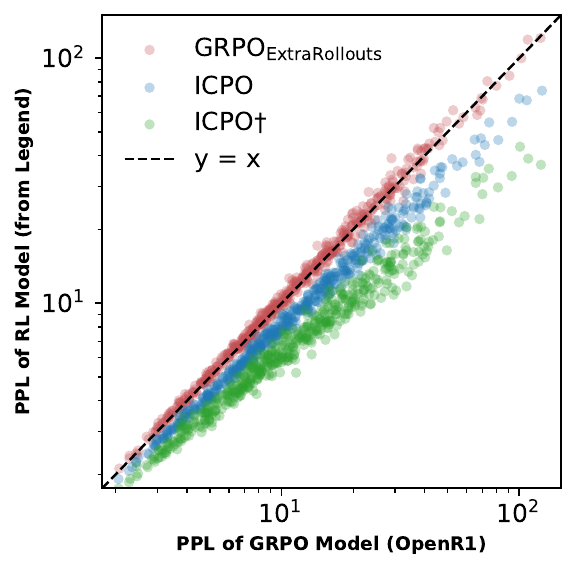}
  \end{subfigure}
  \begin{subfigure}[t]{0.49\linewidth}
    \centering
    \includegraphics[width=\linewidth,height=\linewidth]{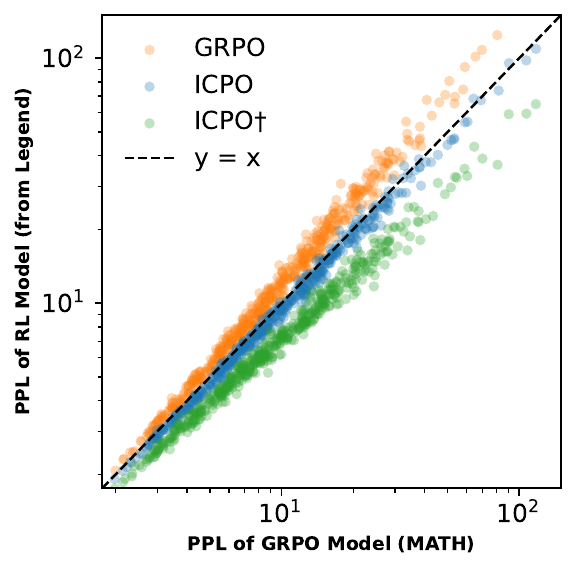}
  \end{subfigure}
  \caption{Perplexity as an instance-level measure of expert-domain distribution shift.}
  \label{fig:analysis_ppl_1.7b}
\end{figure}

\paragraph{Instance-Level.}
Figure~\ref{fig:analysis_ppl_1.7b} presents per-instance perplexity, where each point corresponds to an individual test case. Points lying below the diagonal $y=x$ indicate that the target method yields lower perplexity than GRPO on the same instance, suggesting a shift of the predictive distribution toward the expert domain.
Simply increasing the number of on-policy rollouts yields no deviation from the GRPO baseline, indicating that rollout amplification alone does not alter the underlying policy distribution. In contrast, ICPO induces a clear shift toward expert-like reasoning patterns, and incorporating expert-bonus RS further strengthens it.

\begin{figure}[t]
  \centering
  \includegraphics[
    width=\columnwidth,
    trim=0pt 15pt 0pt 5pt,
    clip
  ]{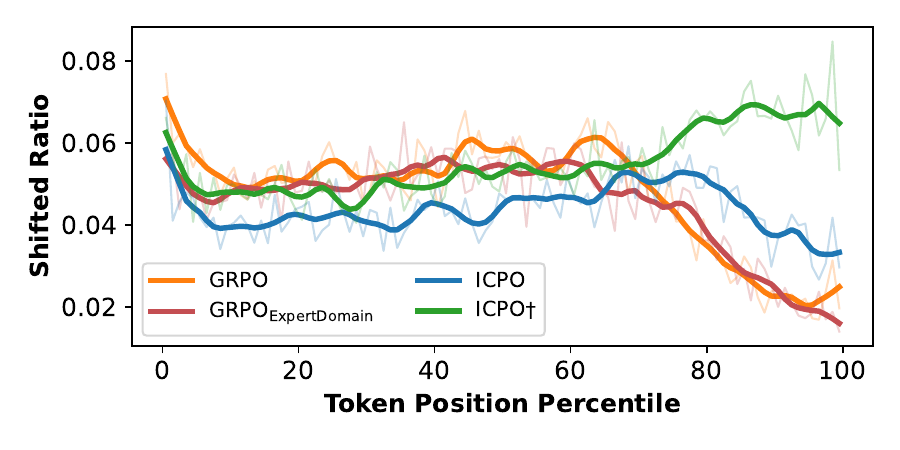}
  \caption{Token-rank shift as a token-level measure of expert-domain distribution shift.}
  \label{fig:analysis_pos_shift_8b}
\end{figure}

\paragraph{Token-Level.}
Figure~\ref{fig:analysis_pos_shift_8b} presents token-rank shift \cite{lin2023tokenrankshift} based on percentile positions (details are in Appendix~\ref{sec:token_rank_shift}). For 0\%–60\%, all methods exhibit similar stable trends. Beyond the 60\% percentile, the differences become pronounced: GRPO shows a steep decline, with the model reverting to high-probability outputs of the base policy in later generations, indicating exploration collapse.
In contrast, ICPO exhibits a slower drop in token shift ratio during later stages, as IEF continuously provides directional guidance toward the expert domain. Notably, ICPO$\dagger$ benefits from the expert-bonus RS, which serves as an effective regularizer, resulting in the largest shift ratio in later generations. As a result, it is more likely to explore solutions that are different from the base policy.

Overall, the results demonstrate that ICPO effectively enables LRM to “\textit{\textbf{think outside the policy}}.”

\section{Conclusion}
We have presented ICPO, a unified RLVR framework that enhances reasoning without relying on external expert models. Leveraging the inherent ICL capability of LRMs, ICPO introduces mixed-policy GRPO with IEF, which constructs expert-conditioned rollouts from existing datasets, improving data utilization and expanding exploration beyond the current policy distribution.
To ensure stable optimization, ICPO further integrates ERRS to eliminate noisy off-policy trajectories and adopts RS to facilitate a smooth transition from expert-guided imitation to autonomous optimization.
Experiments show that ICPO consistently improves RL performance, highlighting its promise as a scalable and general post-training paradigm for LRMs.

\section*{Limitations}
Our experiments focus primarily on mathematical reasoning and OOD reasoning benchmarks under the RLVR framework, including multi-domain knowledge reasoning (MMLU), open-domain QA (GPQA), and scientific reasoning tasks (ARC). Extending ICPO to other domains, such as code generation, may require task-specific adaptations, especially in reward design and evaluation. We leave a systematic investigation of these broader applications to future work.

\section*{Acknowledgments}
This work is supported by the National Natural
Science Foundation of China (62076008).

\section*{Ethics Statement}
\paragraph{Use of AI Assistants.}
We have employed ChatGPT as a writing assistant, primarily for polishing the text after the initial composition.
We certify that any use of AI tools, including ChatGPT, was strictly limited to linguistic refinement such as improving grammar, clarity, and style. All substantive ideas, analyses, and arguments presented in this work originate from the authors or from properly cited prior research.

\paragraph{Computational Budget.}
All our experiments are conducted on a machine with CentOS 8, 384 AMD$^\circledR$ EPYC\texttrademark{} 9K84 96-Core Processor CPUs and 2.2TiB memory. We use 8$\times$ NVIDIA H20 GPUs for all the experiments. The training of ICPO/ICPO$\dagger$ takes around 3 and 7 days for the 1.5B and 8B models, respectively.

\paragraph{Reproducibility.}
Our work is reproducible because we have provided our source code and implementation details.

\paragraph{Potential Risks.}
To the best of our knowledge, there are no potential risks concerning our work.

\paragraph{Licenses.}
The licenses of the scientific artifacts we use are
shown in Table~\ref{tab:license}.
\begin{table}[htbp]
\small
\centering
\setlength{\tabcolsep}{5pt}
\begin{tabular}{lll}
\toprule
\textbf{Category} & \textbf{Artifact} & \textbf{License} \\
\midrule

\multirow{2}{*}{\textbf{Model}}
 & Qwen3 Models     & Apache-2.0 \\
 & Qwen2.5 Models   & Apache-2.0 \\

\midrule
\multirow{3}{*}{\textbf{Framework}}
& LIMO   & MIT \\
 & VERL   & Apache-2.0 \\
 & Math-Verify & Apache-2.0 \\
\midrule
\multirow{10}{*}{\textbf{Dataset}}
 & Skywork-OR1-RL-Data      & Apache-2.0 \\
 & Openr1-Math-46k-8192     & MIT \\
 & MathInstruct             & MIT \\
 & MATH             & MIT \\
 & MATH-500         & MIT \\
 & Minerva          & MIT \\
 & OlympiadBench    & MIT \\
 & AMC23           & N/A \\
 & AIME24          & N/A \\
 & AIME25          & N/A \\
 \midrule
 \multirow{1}{*}{\textbf{Methods}}
 & LUFFY           & N/A \\

\bottomrule
\end{tabular}
\caption{Licenses of scientific artifacts used in this work.}
\label{tab:license}
\end{table}


\bibliography{custom}

\appendix
\onecolumn
\section{Prompt Template}
\label{sec:prompt}
Here we provide the detailed prompt formats used for our experiments.
\begin{figure}[H]
\centering
\begin{subfigure}[t]{\textwidth}
\centering
\begin{tcolorbox}[
    title=Prompt for Qwen3,
    colframe=colframecolor,
    colback=colbackcolor,
]
\setlength{\baselineskip}{0.9\baselineskip}

\textcolor{gray}{\textit{\# Prompt for RL and 0-shot Inference}}\\
\texttt{<|im\_start|>user}\\
\texttt{\{QUESTION\}} Let's think step by step and output the final answer within \textbackslash boxed\{\}\\
\texttt{<|im\_end|>}

\texttt{<|im\_start|>assistant}
\tcblower
\textcolor{gray}{\textit{\# Prompt for few-shot Inference}}\\
\texttt{<|im\_start|>user}\\
Question: \texttt{\{QUESTION\}} Let's think step by step and output the final answer within \textbackslash boxed\{\}\\
Answer: \texttt{\{ANSWER\}} \textbackslash n \textbackslash n \textbackslash n \\
...\\
Question: \texttt{\{QUESTION\}} Let's think step by step and output the final answer within \textbackslash boxed\{\}\\
Answer: \texttt{<|im\_end|>}

\texttt{<|im\_start|>assistant}
\end{tcolorbox}
\label{fig:qwen3prompt}
\end{subfigure}

\vspace{6pt}

\begin{subfigure}[t]{\textwidth}
\centering
\begin{tcolorbox}[
    title=Prompt for Qwen2.5,
    colframe=colframecolor,
    colback=colbackcolor,
]
\setlength{\baselineskip}{0.9\baselineskip}

\texttt{SYSTEM\_PROMPT}$=$Your task is to follow a systematic, thorough reasoning process before providing the final solution. This involves analyzing, summarizing, exploring, reassessing, and refining your thought process through multiple iterations. Structure your response into two sections: \textit{Thought} and \textit{Solution}. In the Thought section, present your reasoning using the format: ``\texttt{<think>\textbackslash nthoughts </think>\textbackslash n}''. Each thought should include detailed analysis, brainstorming, verification, and refinement of ideas. After ``\texttt{</think>\textbackslash n}'' in the Solution section, provide the final, logical, and accurate answer, clearly derived from the exploration in the Thought section. If applicable, include the answer in \texttt{\textbackslash boxed\{\}} for closed-form results such as multiple-choice or mathematical solutions.\\[4pt]
\textcolor{gray}{\textit{\# Prompt for RL and 0-shot Inference}}\\
\texttt{\{SYSTEM\_PROMPT\}}\\
\textbf{User:}\\
Question: \texttt{\{QUESTION\}}\\
Answer: \\[1pt]
\textbf{Assistant:} <think>

\tcblower

\textcolor{gray}{\textit{\# Prompt for few-shot Inference}}\\
\texttt{\{SYSTEM\_PROMPT\}}\\
\textbf{User:}\\
Question: \texttt{\{QUESTION\}}\\
Answer: \texttt{\{ANSWER\} \textbackslash n \textbackslash n \textbackslash n}\\
\texttt{...}\\
Question: \texttt{\{QUESTION\}}\\
Answer: \\[1pt]
\textbf{Assistant:} <think>
\end{tcolorbox}
\label{fig:qwen25prompt}
\end{subfigure}

\caption{Prompt formats used for RL training, zero-shot inference, and few-shot inference.}
\label{fig:prompt_formats}
\end{figure}

\twocolumn

\section{Experimental Details}

\subsection{Training Settings}
\label{sec:train_setting}
For RL finetuning, we use the widely adopted GRPO algorithm built on \textit{VERL} \cite{2025verlframework} framework. The full hyperparameters used in our training are listed in Table~\ref{tab:hparam}. Specifically, for rollout generation, we use a temperature of 1.0, and rewards are computed using \textit{Math-Verify}. All models are trained for $T=400$ optimization steps, and we report results using the final checkpoint.
\begin{table}[htbp]
  \centering
    \begin{tabular}{lc}
    \toprule
    \textbf{Hyper-parameter} & \textbf{Value} \\
    \midrule
    Learning Rate & 1e-6 \\
    Totel Steps & 400 \\
    Batch Size & 128 \\
    Mini Batch Size & 64 \\
    KL Loss Coefficient & 0.0 \\
    Clip Ratio & 0.2 \\
    Temperature & 1.0 \\
    Total Number of Rollouts & 8 \\
    Maximum Prompt Length & 4096 \\
    Maximum Response Length & 8192 \\
    \bottomrule
    \end{tabular}
  \caption{Full hyper-parameters for training.}
  \label{tab:hparam}
\end{table}

\subsection{Evaluation Settings}
\label{sec:eval_setting}
For main experiments, the AIME24, AIME25, and AMC23 benchmark, which have relatively small test sets, we report \textbf{\textit{Avg@32}} (for \textit{\textbf{Pass@1}}), while for the other benchmarks we report \textbf{\textit{Pass@1}}. As a complementary analysis, we also report \textbf{\textit{Pass@32}} results in Appendix~\ref{sec:pass32}. All evaluations are conducted using the \textit{LIMO} framework~\cite{ye2025limo}. During inference, we follow \citet{luffy2025} and set the generation temperature to 0.6.
For ICL evaluation, we randomly sample demonstrations from the \textit{MATH}~\cite{math2021} training set using 5 different random seeds, and report the average performance across them.

\subsection{Baseline Methods}
\label{sec:baselines}
We benchmark ICPO against the following baselines using \textit{Qwen2.5-Math-7B} \cite{yang2024qwen25mathtechnicalreportmathematical} as the base model.
We compare our method with three lines of work, which is previous RLVR methods, SFT and RL baseline, and methods combining SFT and RL that is comparable with our ICPO.

\paragraph{Previous RLVR Methods.}
(1) \textbf{SimpleRL-Zero} \cite{zeng2025simplerl}, which applies GRPO to approximately 24k mathematical samples from \textit{GSM8K} \cite{cobbe2021trainingverifierssolvemath} and \textit{MATH} dataset \cite{math2021};
(2) \textbf{PRIME-Zero} \cite{cui2025PRIMEzero}, which conducts policy rollouts on 150k \textit{NuminaMath} queries with implicit process rewards;
(3) \textbf{OpenReasoner-Zero} \cite{hu2025openreasonerzero}, a PPO-based approach trained on 129k multi-source samples;
and (4) \textbf{Oat-Zero} \cite{liu2025oatzero}, which removes the standard deviation
in GRPO advantage calculation, and is trained on the \textit{MATH} dataset.

\paragraph{SFT and RL.} Here we consider three kind of methods:
(1) \textbf{RL$_\text{GRPO}$}, which is train on-policy within RLVR paradigm using GRPO with the same reward and data as ICPO;
(2) \textbf{SFT }, where the model is supervised on the same prompts and reasoning traces as LUFFY~\cite{luffy2025};
(3) \textbf{SFT+RL}, which is a two-stage training that continues RL training after SFT.

\paragraph{Comparable Baseline Methods.}
For methods that integrate supervised fine-tuning (SFT) with reinforcement learning (RL), we evaluate three representative approaches:  
(1) \textbf{LUFFY}~\cite{luffy2025}, a mixed-policy GRPO method that incorporates advanced LRM reasoning trajectories as off-policy rollouts;  
(2) \textbf{ReLIFT}~\cite{ma2025ReLIFT}, which alternates standard RLVR updates with targeted supervised fine-tuning on the hardest online-collected questions, enabling the model to acquire new reasoning skills that pure RL cannot provide;  
(3) \textbf{Prefix-RFT}~\cite{huang2025RLPrefixSampling}, which samples a prefix from advanced LRM reasoning trajectories and reinforces the model’s continuation, mixing these hybrid sequences with online rollouts under the RFT objective to unify SFT-style imitation and RFT-style exploration.

All these methods rely on advanced LRM-generated reasoning trajectories from the training set as demonstrations, whereas ICPO leverages few-shot demonstrations retrieved from an existing external dataset—independent of the training set—and uses them purely as in-context prompts to steer the rollout distribution on-policy.

\subsection{Data Statistics}
\label{sec:data_statistic}
\begin{figure}[htbp]
  \centering
  \begin{subfigure}[t]{0.49\linewidth}
    \centering
    \includegraphics[width=\linewidth]{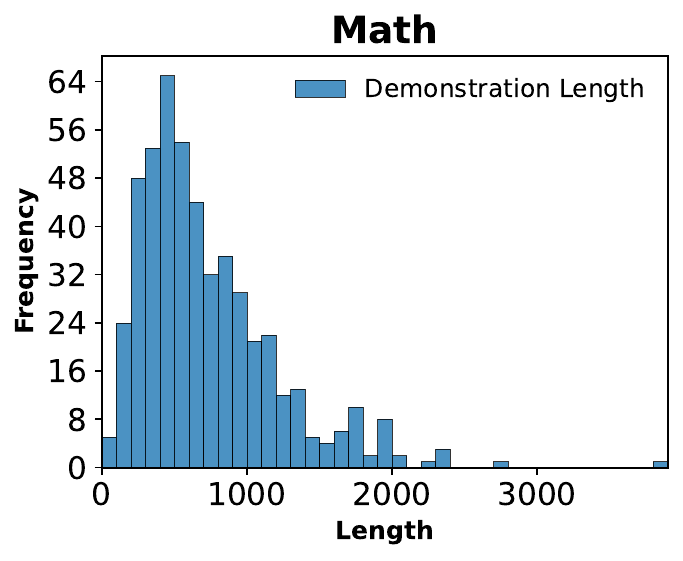}
  \end{subfigure}
  \begin{subfigure}[t]{0.49\linewidth}
    \centering
    \includegraphics[width=\linewidth]{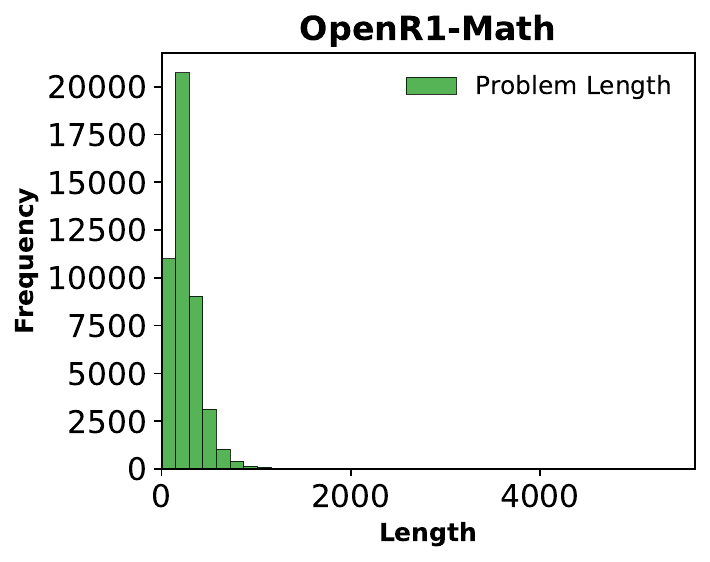}
  \end{subfigure}
  \caption{Length distribution of expert demonstrations (left) and training prompts (right).}
  \label{fig:data_stat}
\end{figure}

Figure~\ref{fig:data_stat} presents the length distribution of the demonstrations 
(question + answer) and the training prompts. We observe that the demonstrations 
are relatively short, ensuring that they do not cause input truncation or the loss 
of important information. Compared to approaches that append the model’s generated 
reasoning traces as additional context, our overall input length remains much 
shorter and thus more efficient.

\section{Demonstration Selection Strategies}
\label{sec:demo_selection}

\begin{figure*}[htbp]
  \centering
  \begin{subfigure}[t]{0.24\linewidth}
    \centering
    \includegraphics[width=\linewidth]{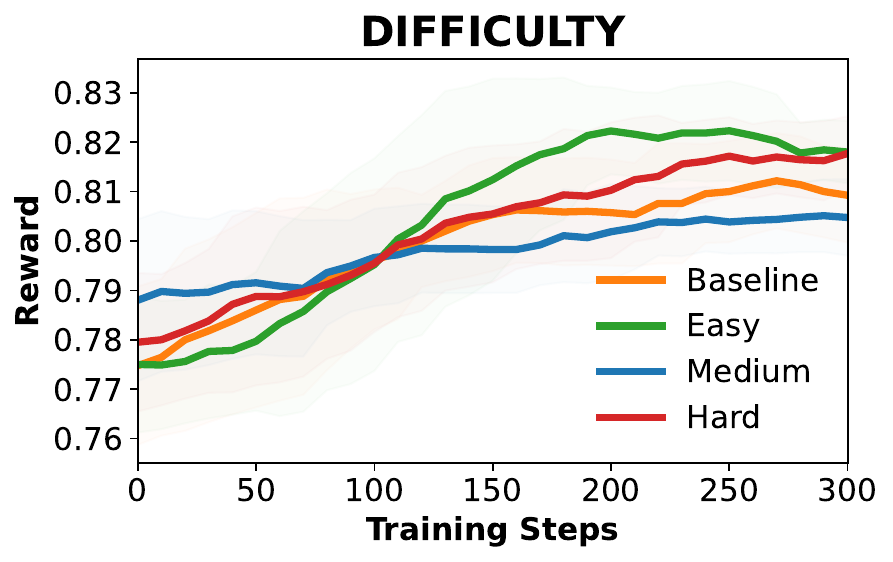}
  \end{subfigure}
  \begin{subfigure}[t]{0.24\linewidth}
    \centering
    \includegraphics[width=\linewidth]{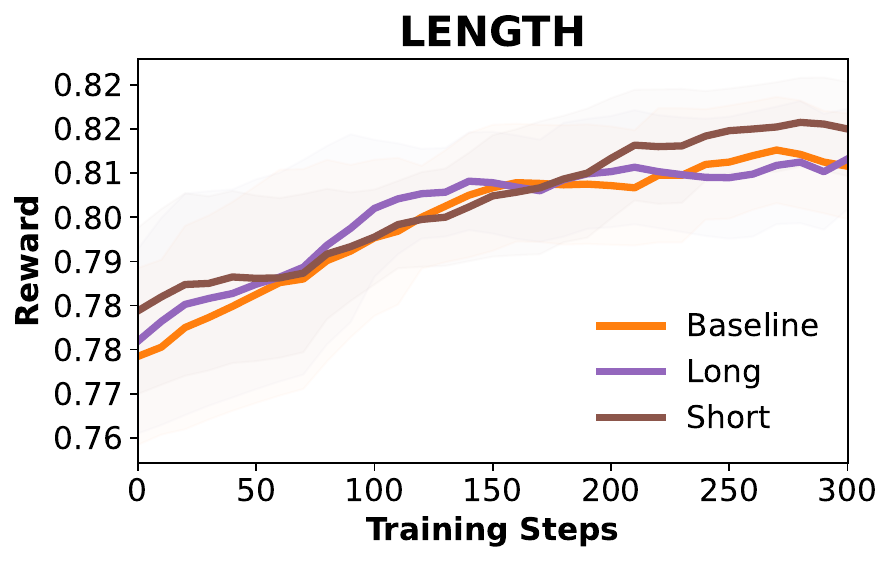}
  \end{subfigure}
  \begin{subfigure}[t]{0.24\linewidth}
    \centering
    \includegraphics[width=\linewidth]{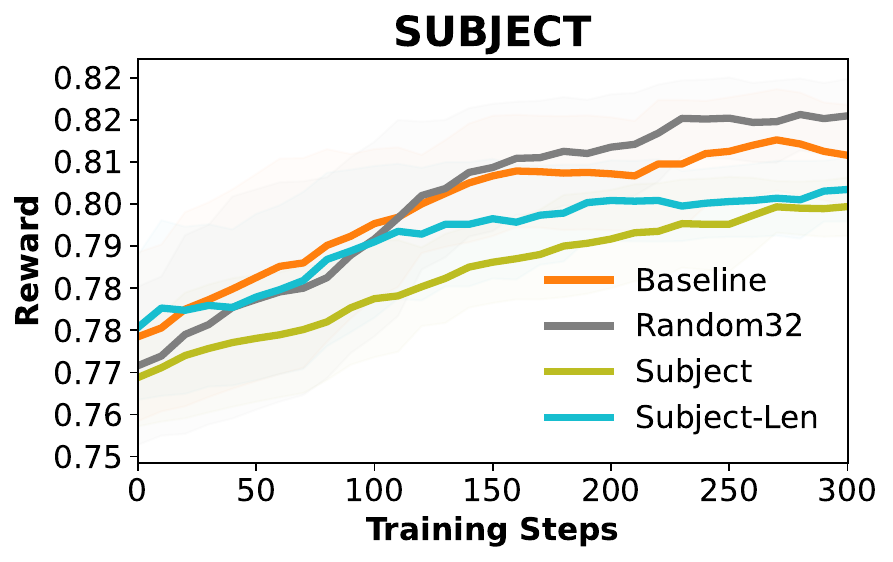}
  \end{subfigure}
  \begin{subfigure}[t]{0.24\linewidth}
    \centering
    \includegraphics[width=\linewidth]{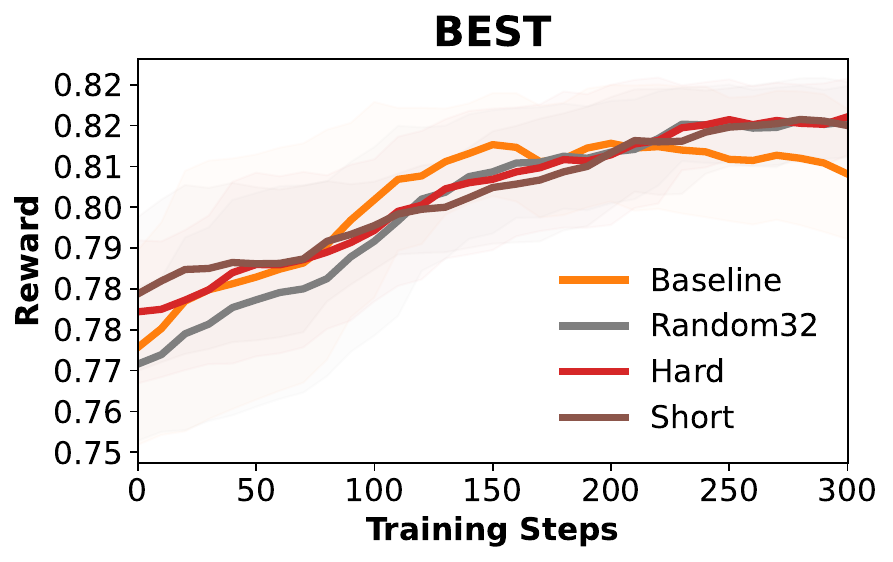}
  \end{subfigure}
  \caption{Rewards on the \textit{MATH-500} dataset for different heuristic demonstration-selection strategies using \textit{Qwen3-1.7B}. The \textit{\textbf{Hard}}, \textit{\textbf{Short}}, and \textit{\textbf{Random32}} demonstration strategies achieve comparable performance and all outperform the GRPO baseline.}
  \label{fig:demo_selection}
\end{figure*}

\begin{figure}[htbp]
  \centering
  \begin{subfigure}[t]{0.49\linewidth}
    \centering
    \includegraphics[width=\linewidth]{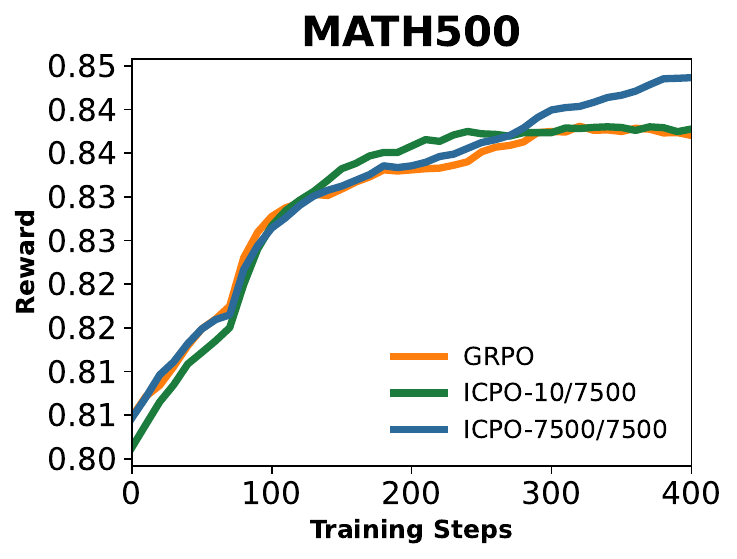}
  \end{subfigure}
  \begin{subfigure}[t]{0.49\linewidth}
    \centering
    \includegraphics[width=\linewidth]{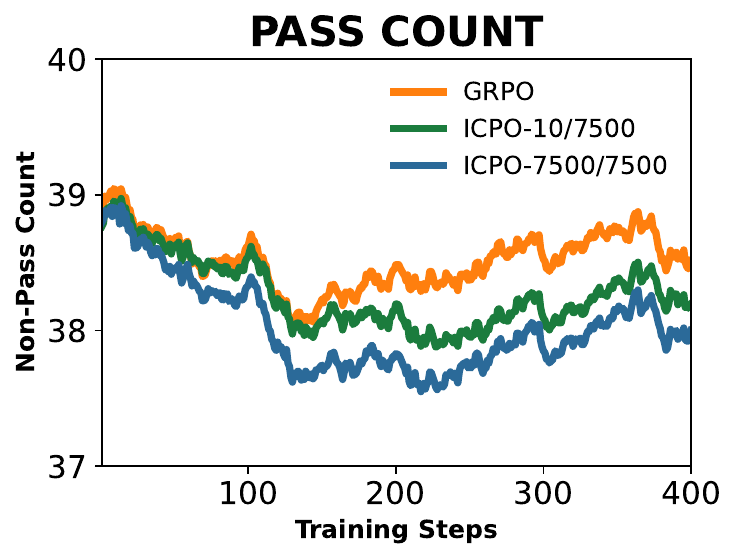}
  \end{subfigure}
  \caption{Comparison of ICPO training using different demonstration pool sizes on \textit{Qwen3-8B}. The left figure reports the reward and training dynamics on the MATH-500 benchmark, while the right figure presents the number of non-pass cases on the training set.}
  \label{fig:pool_size}
\end{figure}

\paragraph{Effects of Heuristic Demonstration Selection Strategies.}
We investigate several heuristic strategies for selecting demonstrations within ICPO. Experiments are conducted on the \textit{Skywork-OR1-RL-Data} \cite{he2025skywork} dataset using \textit{Qwen3-1.7B}, with all demonstrations sourced from the training split of \textit{MATH}~\cite{math2021} as our main setup.
We evaluate the following heuristic selection methods:

\begin{itemize}
    \item \textbf{Difficulty-based selection.} 
    We sample each MATH training instance three times: examples solved correctly in all three attempts are labeled \textit{Easy}, those answered incorrectly three times are labeled \textit{Hard}, and the remaining examples are categorized as \textit{Medium}. During training, each prompt is matched with a randomly selected demonstration from its corresponding difficulty group.
    \item \textbf{Length-based selection.}
    We partition both the demonstration pool and the training set into two equal-sized buckets based on response length. For each training sample, a demonstration is randomly selected from the bucket with the same length range.
    \item \textbf{Subject-based selection.}
    We group both the demonstration data and the training data by their problem category (subject). Each training instance is paired with a randomly selected example from the same subject as its 1-shot demonstration.
    \item \textbf{Random.}
    We randomly sample 32 examples from the entire demonstration pool and, for each training instance at every step, randomly select one of these 32 as the demonstration.
\end{itemize}

The results are shown in Figure~\ref{fig:demo_selection}. Among all heuristic strategies, the \textbf{\textit{Hard}}, \textbf{\textit{Short}}, and \textbf{\textit{Random32}} variants achieve comparable performance, and all surpass the GRPO baseline.
Given that difficulty-based and length-based methods require additional sampling or preprocessing cost, we adopt the simplest and most cost-efficient option, namely \textbf{\textit{Random}} demonstration selection, as our default strategy.

\paragraph{Effects of Demonstration Pool Size.}
We further investigate how the size of the demonstration pool available to each batch affects ICPO training using \textit{Qwen3-8B}, with the results shown in Figure~\ref{fig:pool_size}.
 We compare two configurations. The first sets the demo pool size to 10 (denoted as \textit{ICPO-10/7500}): for each batch, we randomly sample 10 demonstrations, and each instance within the batch randomly selects one of these as its 1-shot demonstration. This design aims to prevent excessive steering directions within a batch, which may destabilize training. 
The second configuration sets the demonstration pool size equal to the full demonstration set (denoted as \textit{ICPO-7500/7500}), where each training instance samples its 1-shot demonstration from the entire pool of 7,500 examples, thereby increasing exploration during training.

As shown in the results, \textit{ICPO-10/7500} yields more stable and stronger early-stage performance but restricts exploration in later stages. In contrast, \textit{ICPO-7500/7500} demonstrates greater potential in the later phase, achieving higher rewards on the expert domain (MATH-500) and resolving more non-pass cases. Based on these observations, we adopt random selection with a full demonstration pool size of 7,500 as the default configuration for our main experiments.

\section{Results and Analysis}

\begin{figure*}[t]
\centering 
\begin{subfigure}[t]{0.32\linewidth}
\includegraphics[width=\linewidth]{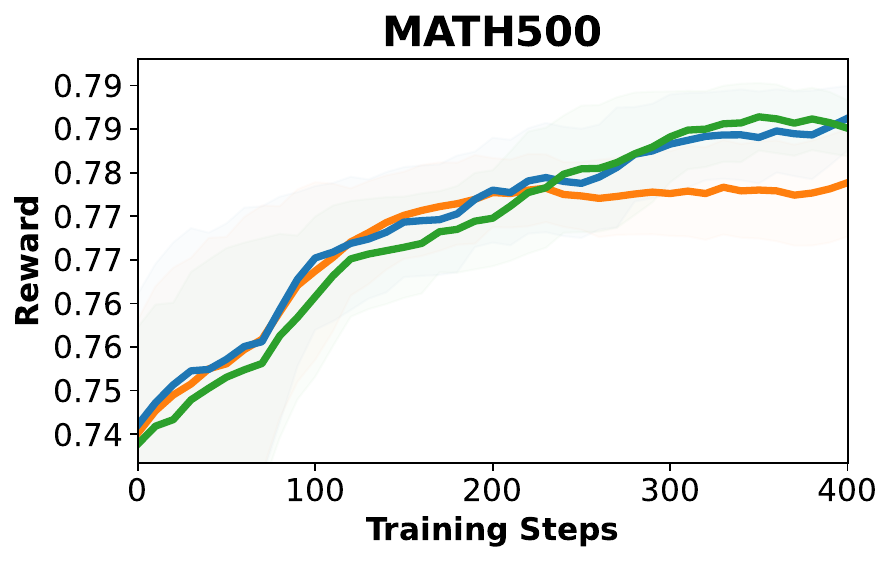}
\caption{Expert Domain Reward}
\end{subfigure} %
\begin{subfigure}[t]{0.32\linewidth}
\includegraphics[width=\linewidth]{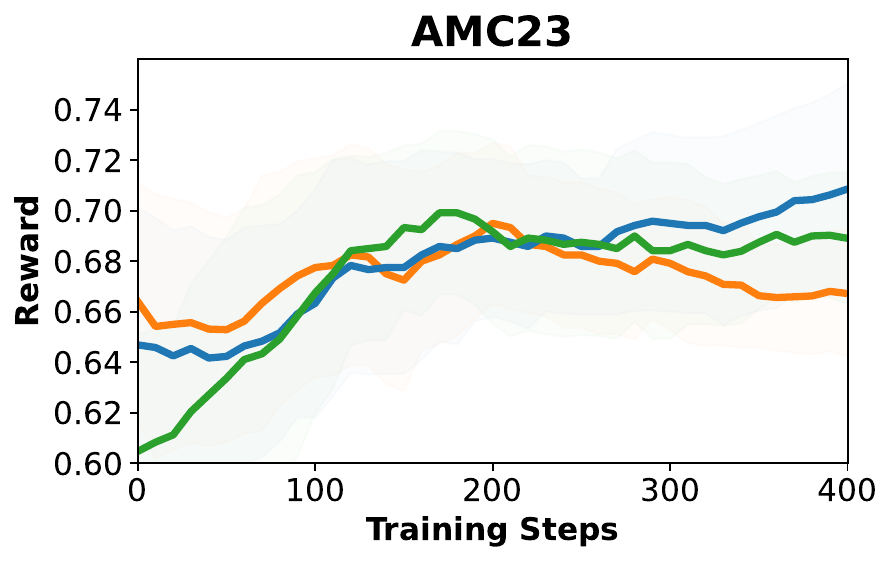}
\caption{In-Distribution Reward}
\end{subfigure}
\begin{subfigure}[t]{0.32\linewidth}
\centering
\includegraphics[width=\linewidth]{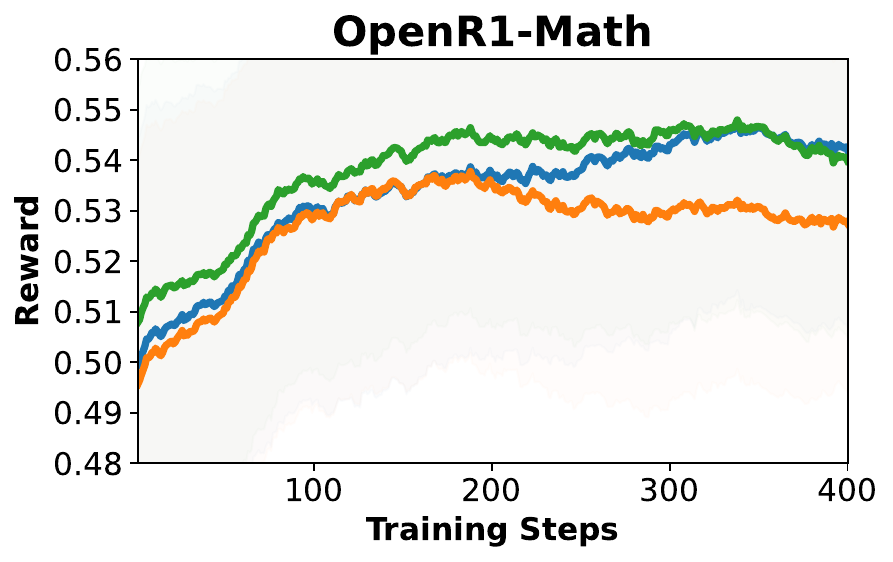}
\caption{Training Reward}
\end{subfigure}
\caption{Reward curves of Qwen3-1.7B over training steps across test and train sets.} \label{fig:reward_curves_1_7B}
\end{figure*}

\subsection{Pass@32 Performance}
\label{sec:pass32}
We report \textit{Pass@32} results to further analyze the exploration behavior of our ICPO.
As shown in Table~\ref{tab:qwen3_8b_pass32results}, simply increasing the number of rollouts ($\text{GRPO}_{\text{ExtraRollouts}}$) or directly training on expert-domain data ($\text{GRPO}_{\text{ExpertDomain}}$) yields only limited or inconsistent improvements in \textit{Pass@32} across benchmarks. In contrast, ICPO achieves higher scores across both \textit{Qwen3-8B-Base} and \textit{Qwen3-8B}, indicating that IEF and RS during RL training effectively guides exploration toward expert-aligned regions of the solution space, enlarging the set of reachable correct solutions rather than merely sharpening the policy around a single mode. This behavior aligns with our motivation of leveraging LRM's inherent ICL ability to enhance exploration without relying on advanced LRMs.

\begin{table*}[htbp]
\centering
\small
\setlength{\tabcolsep}{4pt}
\begin{tabular}{lcccccccc}
\toprule
\textbf{Method} 
& \textbf{AIME24} 
& \textbf{AIME25} 
& \textbf{AMC23} 
& \textbf{MATH-500} 
& \textbf{Minerva} 
& \textbf{Olympiad} 
& \textbf{Average} \\

\midrule
\textbf{\textit{Qwen3-8B-Base}} 
& 46.7 & 40.0 & 85.0 & 94.2 & \textbf{65.4} & 68.3 & 66.6 \\
\midrule
\; GRPO
& 53.3 & 43.3 & 90.0 & \textbf{96.2} & 58.8 & 72.7 & 69.1 \\

\; $\text{GRPO}_{\text{ExtraRollouts}}$
& 50.0 & 46.7 & 92.5 & 95.4 & 64.3 & \textbf{73.8} & 70.4 \\

\; $\text{GRPO}_{\text{ExpertDomain}}$
& 46.7 & 26.7 & 90.0 & 92.0 & 61.0 & 64.2 & 63.4 \\

\midrule

\; ICPO (Ours)
& \textbf{60.0} & \textbf{60.0} & \textbf{95.0} & 95.8 & 63.6 & 73.0 & \textbf{74.6} \\

\; ICPO$\dagger$ (Ours)
& 56.7 & 40.0 & \textbf{95.0} & \textbf{96.2} & 62.1 & 72.7 & 70.5 \\

\midrule
\midrule
\textbf{\textit{Qwen3-8B}} 
& 56.7 & 36.7 & 85.0 & 93.2 & 57.7 & 57.5 & 64.5 \\

\midrule

\; GRPO
& 80.0 & 60.0 & \textbf{95.0} & 97.0 & 65.1 & 79.0 & 79.3 \\

\; $\text{GRPO}_{\text{ExtraRollouts}}$
& 80.0 & 60.0 & \textbf{95.0} & 97.0 & 62.9 & 79.4 & 79.0 \\

\; $\text{GRPO}_{\text{ExpertDomain}}$
& 83.3 & 66.7 & 92.5 & \textbf{97.8} & 64.7 & 77.6 & 80.4 \\

\midrule

\; ICPO (Ours)
& \textbf{86.7} & 66.7 & \textbf{95.0} & 96.8 & 65.4 & 75.0 & 80.9 \\

\; ICPO$\dagger$ (Ours)
& 80.0 & \textbf{70.0} & \textbf{95.0} & \textbf{97.8} & \textbf{67.7} & \textbf{81.6} & \textbf{82.0} \\

\bottomrule
\end{tabular}
\caption{Pass@32 results for Qwen3-8B-Base and Qwen3-8B.}
\label{tab:qwen3_8b_pass32results}
\end{table*}

\subsection{OOD Performance}
\label{sec:qwen25ood}
We evaluate ICPO and the corresponding baselines on out-of-distribution (OOD) benchmarks using \textit{Qwen2.5} models, as shown in Table~\ref{tab:qwen25_in_dis_result} and Table~\ref{tab:qwen25_ood}.
Although \textit{Qwen2.5-Math-7B} performs reasonably well on in-distribution datasets, it fails to outperform other baselines on OOD evaluations. In contrast, stronger base models such as \textit{Qwen3} achieve consistently superior performance on both in-distribution and OOD settings (see Table~\ref{tab:main_results}).

We hypothesize that this discrepancy stems primarily from differences in the base models’ inherent ICL capabilities. Models with stronger in-context learning ability exhibit more robust generalization, and consequently achieve better overall performance across both in-distribution and OOD tasks.

\begin{table*}[t]
\centering
\small
\setlength{\tabcolsep}{8pt}
\begin{tabular}{lcccc}
\toprule
\textbf{Method} 
& \textbf{ARC-C} & \textbf{GPQA-D} & \textbf{MMLU-Pro} 
& \textbf{Average} \\
\midrule
\textbf{\textit{Qwen2.5-Math-7B}}~\cite{yang2025qwen3technicalreport}
& 18.2 & 11.1  & 16.9 & 15.4 \\

\midrule
\multicolumn{5}{l}{\textbf{Previous RLVR Methods}}\\[3pt]
\; SimpleRL-Zero* \cite{zeng2025simplerl}
& 30.2 & 23.2 & 34.5 & 29.3 \\
\; PRIME-Zero* \cite{cui2025PRIMEzero}
& 73.3 & 18.2 & 32.7 & 41.4 \\
\; OpenReasoner-Zero* \cite{hu2025openreasonerzero}
& 66.2 & 29.8 & 58.7 & 51.6 \\
\; Oat-Zero* \cite{liu2025oatzero}
& 70.1 & 23.7 & 41.7 & 45.2 \\

\midrule
\multicolumn{5}{l}{\textbf{SFT and RL}}\\[3pt]
\; SFT~\cite{luffy2025}
& 74.7 & 28.3 & 44.4 & 49.1 \\
\; RL~\cite{grpo2024deepseek}
& 75.9 & 32.8 & 42.6 & 50.4 \\
\; SFT+RL~\cite{grpo2024deepseek}
& 79.2 & 37.9 & 49.6 & 55.5 \\

\midrule
\multicolumn{5}{l}{\textbf{Comparable Baseline Methods}}\\[3pt]
\; ReLIFT \cite{ma2025ReLIFT}
& 74.9 & 40.9 & 51.9 & 55.9 \\
\; LUFFY \cite{luffy2025}
& 80.5 & 39.9 & 53.0 & 57.8 \\
\; Prefix-RFT \cite{huang2025RLPrefixSampling}
& 84.0 & 39.1 & 52.1 & 58.4 \\

\rowcolor{black!5}
\; ICPO (Ours)
& 74.0 & 34.3 & 46.5 & 51.6 \\
\rowcolor{black!5}
\; ICPO$\dagger$ (Ours)
& 77.0 & 33.8 & 47.6 & 52.8 \\

\bottomrule
\end{tabular}
\caption{Evaluation results for Qwen2.5-Math-7B. Methods marked with “*” indicate results reported from their original papers.}
\label{tab:qwen25_ood}
\end{table*}

\subsection{Generalizability Across Models}
\begin{table*}[htbp]
\centering
\small
\setlength{\tabcolsep}{6pt}
\begin{tabular}{lccccccc}
\toprule
\textbf{Method} 
& \textbf{AIME24} 
& \textbf{AIME25} 
& \textbf{AMC23} 
& \textbf{MATH-500} 
& \textbf{Minerva} 
& \textbf{Olympiad} 
& \textbf{Average} \\
\midrule
\textbf{\textit{Qwen3-8B-Base}}
& 10.8 & 10.4 & 47.6 & 67.2 & 32.0 & 34.4 & 33.7 \\
\midrule
\; GRPO
& 22.5 & 21.4 & \textbf{71.8} & 84.6 & 43.8 & 50.1 & 49.0 \\

\; $\text{GRPO}_{\text{ExtraRollouts}}$
& \textbf{26.9} & 20.2 & 67.4 & 85.8 & 46.7 & 51.6 & 49.8 \\

\; $\text{GRPO}_{\text{ExpertDomain}}$
& 22.9 & 12.8 & 60.9 & 80.0 & 47.8 & 42.8 & 44.5 \\

\; LUFFY
& 24.6 & \textbf{22.1} & 68.2 & 87.2 & 47.4 & 52.3 & 50.3 \\

\midrule

\; ICPO
& 26.8 & 21.6 & 69.2 & \textbf{88.6} & 46.3 & \textbf{53.8} & \textbf{51.0} \\

\; ICPO$\dagger$ 
& 25.9 & 16.9 & 67.1 & 86.6 & \textbf{49.3} & 51.3 & 49.5 \\

\bottomrule
\end{tabular}
\caption{Results on mathematical reasoning benchmarks for Qwen3-8B-Base.}
\label{tab:qwen3_8b_base}
\end{table*}
\begin{table*}[htbp]
\centering
\small
\setlength{\tabcolsep}{4pt}
\begin{tabular}{lcccccccc}
\toprule
\textbf{Method} 
& \textbf{AIME24} 
& \textbf{AIME25} 
& \textbf{AMC23} 
& \textbf{MATH-500} 
& \textbf{Minerva} 
& \textbf{Olympiad} 
& \textbf{Average} \\
\midrule

\textbf{\textit{LLaMA-3.1-8B-Instruct}}
& \textbf{4.6} & 0.2 & \textbf{20.9} & \textbf{45.4} & \textbf{22.8} & 15.7 & \textbf{18.3} \\

\midrule

\; SFT$^{*}$ 
& 0.5 & 0.1 & 5.4 & 20.2 & 4.0 & 5.3 & 5.9 \\

\; GRPO
& 3.1 & \textbf{1.0} & 10.8 & 28.2 & 17.6 & \textbf{16.2} & 12.8 \\

\; LUFFY$^{*}$ 
& 1.9 & 0.1 & 13.5 & 39.0 & 15.1 & 9.6 & 13.2 \\


\midrule

\; ICPO
& 3.1 & 0.2 & 16.6 & 33.4 & 21.0 & 9.0 & 13.9 \\

\; ICPO$\dagger$ 
& 0.4 & 0.1 & 13.4 & 38.2 & 22.4 & 12.0 & 14.4 \\

\bottomrule
\end{tabular}
\caption{Results on mathematical reasoning benchmarks for LLaMA-3.1-8B \cite{grattafiori2024llama3}. Methods marked with “*” indicate results reported from \citet{luffy2025}.}

\label{tab:llama_result}
\end{table*}

To evaluate the generalizability of our method across various model architectures, we extend ICPO to the base model \textit{Qwen3-8B-Base} and \textit{LLaMA-3.1-8B}. As shown in Table~\ref{tab:qwen3_8b_base} and Table~\ref{tab:llama_result}, ICPO variants surpass SFT, RL baseline and LUFFY~\cite{luffy2025} on six mathematical benchmarks, underscoring its reasoning performance.

\section{Ablation Study}
\label{sec:ablation_appendix}
\subsection{Ablation on Expert Guidance}
\label{sec:ablation_expert_guidance}

\paragraph{Expert-Domain Data.}
We study the generalizability of ICPO under different sources of expert-domain data. In addition to our default setting, we adopt \textit{MathInstruct}\footnote{\scriptsize \href{https://huggingface.co/datasets/TIGER-Lab/MathInstruct}{\texttt{https://huggingface.co/datasets/TIGER-Lab/MathInstruct}}}~\cite{yue2023mammothPoT} as an alternative expert demonstration corpus for ICPO. \textit{MathInstruct} contains approximately 262k instruction–solution pairs. To enable reliable extraction of ground-truth answers for RLVR, we select a subset of 118k examples whose solutions explicitly include the phrase ``The answer is''.

As shown in Table~\ref{tab:ablation_expert_data}, ICPO consistently outperforms GRPO when using either CoT or PoT data as demonstrations for implicit expert guidance. These results suggest that ICPO is robust to the choice of expert data, and highlight its potential to generalize beyond mathematical reasoning to cross-domain settings, such as code-oriented tasks.

\subsection{Ablation on Difficulty Levels}
\paragraph{Train Data.}

We additionally train on a simpler dataset, \textit{Skywork-OR1-RL-Data}\footnote{\scriptsize \href{https://huggingface.co/datasets/Skywork/Skywork-OR1-RL-Data}{\texttt{https://huggingface.co/datasets/Skywork/OR1-RL-Data}}} \cite{he2025skywork}, to verify the generalization ability of our IEF under different reasoning conditions. This dataset is annotated with difficulty levels predicted by \textit{DeepSeek-R1-Distill-Qwen-7B} \cite{deepseekR12025},
and we use the subset with difficulty level = 1, which corresponds to the easiest reasoning problems,
as our simplified training corpus for controlled comparison.
As shown in Table~\ref{tab:icpo_skywork_vs_openr1}, results verify that ICPO is consistently beneficial across training datasets of different difficulty levels.
On the simpler Skywork dataset, ICPO improves GRPO by \textbf{+1.9} average points, while on the more challenging OpenR1-Math corpus, the gain further increases to \textbf{+3.4} average points.
This demonstrates that ICPO not only enhances learning on complex reasoning traces but also generalizes effectively to settings with weaker supervision.

\begin{table}[H]
\centering
\small
\begin{threeparttable}
\setlength{\tabcolsep}{4pt}
\begin{tabular}{lccccc}
\toprule
\textbf{Method} & \textbf{MATH} & \textbf{AIME24/25} & \textbf{AMC} & \textbf{Mnrv.} & \textbf{Avg.} \\
\midrule
\multicolumn{6}{l}{\textbf{\textit{OpenR1-Math-220k}}} \\ [3pt]
GRPO & 83.6 & 28.4 / 22.5 & 66.7 & 40.8 & 48.4 \\
ICPO & \textbf{86.8} & \textbf{31.3} / \textbf{26.3} & \textbf{70.4} & \textbf{44.1} & \textbf{51.8} \\
\midrule[0.5pt]
\multicolumn{6}{l}{\textbf{\textit{Skywork-OR1-RL-Data}}} \\ [3pt]
GRPO & 83.0 & 25.1 / 22.2 & 66.9 & 42.3 & 47.9 \\
ICPO & \textbf{86.0} & \textbf{26.8} / \textbf{24.1} & \textbf{69.6} & \textbf{42.6} & \textbf{49.8} \\
\bottomrule
\end{tabular}
\caption{Comparison of ICPO performance on Qwen3-1.7B under two training regimes.}
\label{tab:icpo_skywork_vs_openr1}
\end{threeparttable}
\end{table}

\paragraph{Prompt Groups.}

Table~\ref{tab:ief_ablation} further analyzes the effectiveness of ICPO by grouping prompts according to their initial rollout success rates. We denote prompts with a success rate of \texttt{0.0} as \textit{Non-Pass}, and those with success rates between \texttt{0.0} and \texttt{1.0} as \textit{Some-Pass}. ICPO applies implicit expert forcing (IEF) across all prompt groups.

We observe that expert guidance consistently improves performance for both \textit{Some-Pass} and \textit{Non-Pass} prompts. Moreover, the impact of IEF differs across prompt types and exhibits complementary benefits. For \textit{Some-Pass} prompts, IEF enriches solution diversity while remaining within expert-aligned regions, thereby mitigating premature policy specialization. For \textit{Non-Pass} prompts, IEF provides reliable expert guidance that enables the policy to solve prompts that cannot be handled by vanilla GRPO.

Taken together, these results demonstrate that ICPO enhances exploration for \textit{Some-Pass} cases while enabling effective learning on challenging \textit{Non-Pass} prompts.

\begin{table}[H]
\centering
\small
\begin{threeparttable}
\setlength{\tabcolsep}{0.5pt}
\begin{tabular}{lccccc}
\toprule
\textbf{Variant} & \textbf{MATH} & \textbf{AIME24/25} & \textbf{AMC} & \textbf{Mnrv.} & \textbf{Avg.} \\
\midrule
GRPO & 83.0 & 25.1 / 22.2 & 66.9 & 42.3 & 47.9 \\
\;+IEF on Some-Pass & 82.0 & \textbf{26.9} / \textbf{26.0} & 67.8 & \textbf{43.4} & 49.2 \\
\;+IEF on Non-Pass & 82.4 & 25.4 / 23.3 & 67.2 & 42.3 & 48.1 \\
\midrule
ICPO & \textbf{86.0} & 26.8 / 24.1 & \textbf{69.6} & 42.7 & \textbf{49.8} \\
\bottomrule
\end{tabular}
\caption{Ablation study of IEF on Qwen3-1.7B using the \textit{Skywork} dataset, grouped by rollout success rates.
}
\label{tab:ief_ablation}
\end{threeparttable}
\end{table}

\section{Token Rank Shift Analysis.}
\label{sec:token_rank_shift}
To quantify how different training methods alter model behavior at a fine-grained level, we follow \citet{lin2023tokenrankshift} and adopt \emph{token-rank shift} as a token-level diagnostic metric.
The core idea is to measure how unlikely the generated tokens of an aligned model are under the \emph{base model} distribution.

Given a generated trajectory produced by an aligned policy, we replay the same trajectory under the base model and examine, at each generation step, how the aligned token is ranked according to the base model’s conditional distribution.
Let \( x_{<t} \) denote the prefix up to position \( t-1 \), and let \( y_t \) be the token generated by the aligned model at step \( t \).
We compute the base model logits \( \ell^{\text{base}}(\cdot \mid x_{<t}) \) and define the \emph{base-rank} of \( y_t \) as:
\begin{equation} \small
\mathrm{rank}(y_t)
\;=\;
1 + \left|\left\{ v \;\middle|\; 
\ell^{\text{base}}(v \mid x_{<t}) > \ell^{\text{base}}(y_t \mid x_{<t})
\right\}\right|.
\label{eq:tokenrankshift}
\end{equation}
A rank of \(1\) indicates that the aligned token coincides with the base model’s top-1 prediction, while larger ranks indicate increasing deviation from the base policy’s high-probability region.

For interpretability, we further categorize token ranks into three regimes: (1) \textbf{Unshifted}: \( \mathrm{rank}(y_t) = 1 \); (2) \textbf{Marginally shifted}: \( 1 < \mathrm{rank}(y_t) \le k \); (3) \textbf{Shifted}: \( \mathrm{rank}(y_t) > k \),
where \( k \) is a small threshold (e.g., \( k=3 \)).

We define the \emph{shifted ratio} as the fraction of tokens whose base-rank exceeds $k=3$, which serves as a proxy for how frequently the aligned policy samples tokens that would be unlikely under the base model distribution.

Due to the varying lengths of reasoning trajectories, analyzing token rank shift using absolute token positions can be misleading.
Instead, we normalize token positions by trajectory length and compute statistics over length percentiles.
In Figure~\ref{fig:analysis_pos_shift_8b}, we discretize the normalized positions into 100 bins and aggregate token shift metrics within each bin. By aggregating these statistics across token positions, token rank shift enables a fine-grained analysis of \emph{when} and \emph{to what extent} the aligned policy departs from the base policy during generation.

\end{document}